# Investigating Deep Learning Models for Ejection Fraction Estimation from Echocardiography Videos

Shravan Saranyan
Branham High School

Pramit Saha
University of Oxford

## Abstract

Left ventricular ejection fraction (LVEF) is a key indicator of cardiac function and plays a central role in the diagnosis and management of cardiovascular disease. Echocardiography, as a readily accessible and non-invasive imaging modality, is widely used in clinical practice to estimate LVEF. However, manual assessment of cardiac function from echocardiograms is time-consuming and subject to considerable inter-observer variability. Deep learning approaches offer a promising alternative, with the potential to achieve performance comparable to that of experienced human experts. In this study, we investigate the effectiveness of several deep learning architectures for LVEF estimation from echocardiography videos, including 3D Inception, two-stream, and CNN–RNN models. We systematically evaluate architectural modifications and fusion strategies to identify configurations that maximize prediction accuracy. Models were trained and evaluated on the EchoNet-Dynamic dataset, comprising 10,030 echocardiogram videos. Our results demonstrate that modified 3D Inception architectures achieve the best overall performance, with a root mean squared error (RMSE) of 6.79%. Across architectures, we observe a tendency toward overfitting, with smaller and simpler models generally exhibiting improved generalization. Model performance was also found to be highly sensitive to hyperparameter choices, particularly convolutional kernel sizes and normalization strategies. While this study focuses on echocardiography-based LVEF estimation, the insights gained regarding architectural design and training strategies may be applicable to a broader range of medical and non-medical video analysis tasks.

## Keywords
Echocardiography; Left Ventricular Ejection Fraction; Deep Learning; 3D Convolutional Neural Networks; Video Analysis; Automated Cardiac Function Assessment; AI in Healthcare

## Introduction

Heart disease remains the leading cause of mortality worldwide. In 2023 alone, 910,032 deaths in the United States were attributed to cardiovascular disease, accounting for approximately one in every three deaths [1]. Cardiovascular disease encompasses a broad range of conditions, including coronary artery disease, vascular disease, arrhythmias, and congenital heart disease (CHD), either in isolation or in combination. These conditions, together with comorbidities such as hypertension, diabetes, anemia, and hyperthyroidism, can contribute to the development of heart failure (HF), a chronic syndrome characterized by impaired cardiac function and/or structural abnormalities [2,3].

Although HF can be life-threatening, survival rates improve substantially with early diagnosis, appropriate medical management, and mitigation of modifiable risk factors, including smoking, excessive alcohol consumption, poor diet, and chronic stress [2]. Consequently, accurate and timely assessment of cardiac function is critical for both prognosis and treatment planning.

The left ventricular ejection fraction (LVEF) represents the proportion of blood ejected from the left ventricle with each cardiac contraction and is calculated as the ratio of stroke volume to end-diastolic volume. LVEF is a central indicator of systolic cardiac performance and serves as a primary metric for HF classification. The European Society of Cardiology (ESC) categorizes HF into three phenotypes based on LVEF: HF with reduced ejection fraction (HFrEF, LVEF $< 40\%$), HF with mildly reduced ejection fraction (HFmrEF, LVEF 40–49%), and HF with preserved ejection fraction (HFpEF, LVEF $\geq 50\%$) [4,5]. This distinction is clinically significant, as each phenotype is associated with distinct underlying mechanisms and therapeutic strategies. HFrEF is typically preceded by cardiomyocyte loss and has several effective pharmacological treatments, whereas HFpEF is often driven by chronic comorbidities and currently lacks effective disease-modifying therapies [5]. Both conditions can progress to cardiomyopathy and, ultimately, advanced HF, underscoring the importance of precise LVEF measurement.

Echocardiography, or cardiac ultrasound, is the most widely used imaging modality for assessing LVEF in clinical practice. It provides comprehensive information on cardiac chamber volumes and ventricular systolic and diastolic function, while remaining non-invasive, widely accessible, safe, and cost-effective. Transthoracic echocardiography (TTE) is the standard approach for LVEF estimation [4,6]. Although three-dimensional echocardiography offers the highest accuracy, two-dimensional echocardiography is more commonly available in routine clinical settings. In 2D echocardiography, LVEF is typically computed using the Modified Simpson's method, which estimates ventricular volumes by summing a series of traced disks at end-diastole and end-systole [7]. However, this approach is labor-intensive, requires expert interpretation, and is subject to substantial inter-observer variability [8]. While numerous techniques have been proposed to address these limitations, recent advances in deep learning have introduced promising automated alternatives.

In 2020, Ouyang *et al.* addressed this challenge by introducing EchoNet-Dynamic, a video-based deep learning framework trained on large-scale echocardiography data to directly estimate LVEF from echocardiogram videos [8].

In this work, we systematically investigate deep learning approaches for estimating left ventricular ejection fraction from echocardiography videos. We evaluate and compare multiple architectural paradigms including 3D convolutional networks, two-stream models, and CNN–RNN frameworks using the large-scale EchoNet-Dynamic dataset. Through controlled experiments, we analyze the impact of architectural design choices, fusion strategies, and key hyperparameters on prediction accuracy and generalization performance. By identifying strengths and limitations across model families, this study aims to provide practical insights into the design of robust deep learning systems for automated cardiac function assessment, and to inform future research in video-based medical imaging analysis.

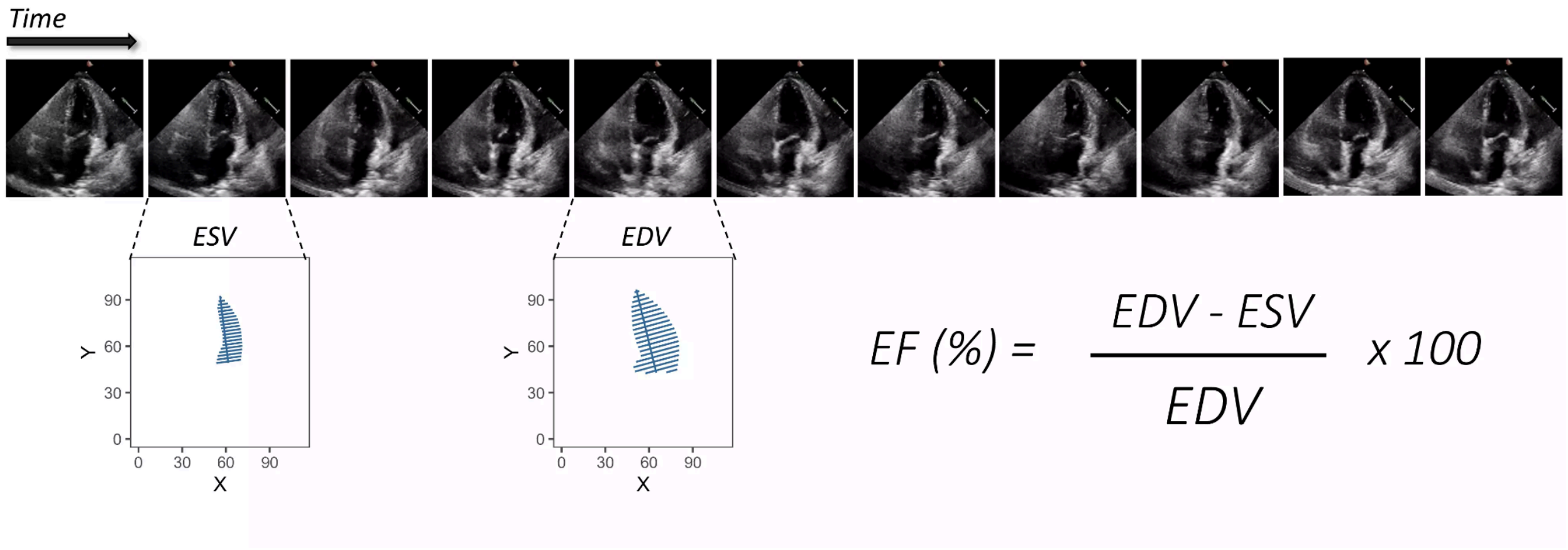

**Figure 1:** Sample frames from the EchoNet-Dynamic dataset [20]. The dataset comprises 10,030 apical four-chamber echocardiography videos that were preprocessed through cropping and masking to remove extraneous text and regions outside the ultrasound scanning sector. The resulting images were subsequently downsampled to a standardized resolution of 112 × 112 pixels using cubic interpolation. The dataset also includes expert-annotated segmentations at end-diastole and end-systole, which are used for ejection fraction (EF) calculation, as illustrated. While these segmentations were not used to train the models presented in this work, they are utilized by selected models described in the *Appendix*.

While numerous deep learning models have been proposed for LVEF estimation, comprehensive benchmarks systematically comparing these architectures for this specific task remain limited. This work aims to address this gap by establishing a comparative benchmark that evaluates the performance of multiple model families, architectural choices, and training parameters for predicting LVEF from echocardiography video clips. We analyze key performance trends to provide insights that may guide future research in LVEF estimation and related medical video analysis problems. Specifically, we evaluate 3D Inception, two-stream, and CNN–RNN architectures, along with several fusion strategies. In total, 51 experiments were conducted to assess the impact of architectural variations, hyperparameter choices, and minor design modifications. Architectures incorporating segmentation heads, as well as additional architectural details, are described in the Appendix. Model performance was evaluated using root mean squared error (RMSE), mean absolute error (MAE), and the coefficient of determination ($R^2$), with prediction-versus-ground-truth and Bland–Altman plots generated to support further analysis.

## Background

Convolutional neural networks (CNNs) are a class of deep neural networks composed of convolutional, pooling, and fully connected layers [9]. Convolutional layers employ learnable kernels (or filters) that are convolved across an input tensor, performing localized operations on each pixel and its neighboring region to produce activation maps [9]. Key hyperparameters governing this process include filter size, stride, and zero-padding. While each filter typically spans the full depth of the input tensor, convolutions are performed across the spatial dimensions. Increasing the filter size expands the network's effective receptive field, *i.e.*, the region of the input that meaningfully influences a given output activation, though this effect saturates beyond a certain scale [38]. The stride determines the step size with which the filter is applied, while zero-padding is commonly used to control the spatial dimensions of the output feature maps [9].

Within a single convolutional layer, multiple filters are applied in parallel, each producing a distinct activation map [11]. These maps are stacked along the channel dimension to form a three-dimensional output tensor, such that each spatial location is represented by a vector of activations whose depth corresponds to the number of filters KKK. Formally, for an input tensor of size $I_H$ x $I_W$ x C and K filters of size $F_H$ x $F_W$ x C, the output activation map takes the shape $O_H$ x $O_W$ x K. The output spatial dimensions are determined as follows [11]:

$$O = \lfloor \frac{I - F + P_{start} + P_{end}}{S} \rfloor + 1$$

Pooling layers perform a downsampling operation to reduce the dimensions of an input. A pooling filter, slides across the input, outputting either the average or maximum value of the region it covers onto the downsampled feature map. Max-pooling is most commonly used, as it preserves prominent features. Average-pooling is generally used to downsample the tensor just before dense layers, improve generalization, and enhance interpretability [12].

Fully connected layers, or Dense layers, are typically placed at the end of the network, and consist of standard feedforward networks. They operate on a flattened input, connecting each input to all output neurons. This layer is used to optimize the final objective, such as producing class scores or outputting a final value for regression [9].

Activation functions introduce non-linearity into the network and are typically applied following dense layers. Traditional functions such as Sigmoid, which maps all values to (0, 1), suffer from the vanishing gradient problem as the model weights become saturated. Today, Rectified Linear Units (ReLU) are the most commonly used function, defined as g(z) = max(0, z). It helps circumvent vanishing gradients by preventing saturation, and is also computationally inexpensive. However, ReLU can cause neurons to become inactive during training if the LR is set too high, variants such as Leaky ReLU and ELU address this issue by allowing a small, non-zero value for negative inputs [9, 13].

Recurrent Neural Networks (RNNs) are a class of neural networks designed for sequential data, possessing hidden states which allow previous outputs to influence further steps [9]. At each step, the RNN combines the current input ($x_t$) and hidden state from the previous step ($a_{t-1}$) to update the state for the next step ($a_t$) and generate an output ($y_t$). This state behaves like a memory, allowing context from previous steps to be utilized [9, 14]. This chain-like structure, inherently related to sequential data, is useful for problems such as NLP and Speech Recognition [9, 14]. However, as sequence length increases, standard RNNs can struggle with long-term dependencies as they encounter vanishing gradients [31]. This issue can be mitigated with gradient clipping, activation functions such as ReLU, and more complex gated units such as GRUs and LSTMs [9].

LSTM (Long Short Term Memory) units use a persistent cell state controlled by three gates to explicitly handle long-term dependencies. The forget gate determines which information is maintained, the input gate determines what information to add to the state, and the output gate determines which information is used to produce the output. GRUs (Gated Recurrent Units) are simpler alternatives, combining the forget and input gates into a single ‘update gate’ and merging the cell and hidden state [9, 14].

Two-Stream Networks for video classification operate on inputs decomposed into their spatial (individual frame appearance) and temporal (motion between frames) components [18]. Simonyan et al. proposed a two-stream video classification model with identical spatial and temporal streams, whose outputs undergo a class score fusion to produce the output. The spatial stream effectively works with still frames, and can be trained with 2D image databases, while the temporal stream operates on optical flow computed between frames, which requires dedicated video databases (UCF-101 and HMDB-51) [18]. As these datasets were small at the time, Simonyan et al. employed multi-task learning to prevent overfitting, using two classification layers for each dataset, computed two loss functions, and backpropagated using the sum of those losses [18].

While increasing model size to improve model performance is common, it often leads to overfitting and requires more computational resources. The authors of Inceptionv1 posited that an optimal, efficient network would employ sparse connections, clustering neurons with highly correlated outputs and connecting them to a single, dedicated unit in the subsequent layer. However, current computing power is highly optimized for dense matrix multiplications, present within fully connected networks. They proposed a compromise by creating “Inception modules”, which mimic the hypothetical sparse network by adding sparsity within the model while utilizing dense matrix multiplications. These modules mimic neuron clustering by performing convolutions at multiple scales (1x1, 3x3, 5x5) simultaneously. The resultant activation maps are then concatenated, extracting features at multiple scales and letting the subsequent layer decide which features are most relevant [15]. To mitigate the computational cost of large filters operating on deep inputs, 1x1 convolutions are strategically used as "bottlenecks" to reduce the channel depth. An additional pooling path is typically included in parallel [15].

Subsequent Inceptionv2 and Inceptionv3 networks further refined this architecture with several key improvements. These included factorizing larger convolution filters into a sequence of smaller filters (e.g. 5x5 into two 3x3) to reduce computational cost while maintaining an equivalent receptive field. To prevent representational bottlenecks, the pooling layer was paired with a simultaneous stride-2 convolution to decrease spatial dimensions while increasing channel depth. Additionally, Batch Normalization and label smoothing were adopted as regularizers [16].

The success of these models for image classification prompted researchers to explore “inflating” them for video classification, creating 3D Inception (I3D) models. 3D convolutions stack contiguous frames into a volume and convolve 3D filters across all three dimensions, allowing temporal information to be treated as a true volumetric signal alongside spatial information [19]. In their implementation, Carreira et al. bootstrapped 2D model weights pre-trained from ImageNet and duplicated them across the temporal dimension. They experimented with various different architectures (Two-Stream, I3D, and Two-Stream I3D) using various video classification test sets for evaluation. The Two-Stream I3D models, which concurrently utilized RGB and Optical Flow channels, consistently outperformed the Two-Stream and I3D models on all test sets [17]. This demonstrated the viability of this architecture for video classification problems.

The Echo-Net Dynamic model predicts LVEF on a beat-by-beat basis using spatiotemporal convolutions with residual connections and semantically segmenting the left ventricle using weak supervision from human tracings, combining these outputs to generate the final prediction [8]. It was the first model to predict LVEF with echocardiogram videos as the sole input, and was able to overcome the time and variability constraints of human assessment while matching or even exceeding the accuracy of human-based approaches. The model was able to achieve an R^2 of 0.81, mean absolute error of 4.1%, and root mean squared error of 5.3%, placing its performance within the range of inter-observer variation [5].

## Materials and Methods

**Dataset and Implementation Details:** All experiments were conducted using the EchoNet-Dynamic dataset [8], which comprises 10,030 labeled echocardiography videos. The dataset includes expert human annotations and is partitioned into training (7,465 videos), validation (1,288 videos), and test (1,277 videos) sets [22]. The videos reflect a wide range of echocardiography acquisition conditions and are accompanied by corresponding ejection fraction (EF) measurements, end-systolic and end-diastolic left ventricular volumes, and expert tracings [20].

While the original EchoNet-Dynamic study employed 32-frame clips and leveraged human tracings for weak supervision, our approach relies exclusively on 28-frame clips and does not use segmentation during training [8]. In both cases, videos were cropped and resized to 112 × 112 pixels, and EF values were used as ground-truth regression targets. The dataset is publicly available on Kaggle under the name heart-database [21].

All experiments were implemented in TensorFlow/Keras and executed on Google Colab, primarily using NVIDIA A100 and L4 High-RAM GPUs [36,40]. To improve training efficiency, raw Kaggle files were preprocessed into TFRecord format. Separate TFRecord pipelines were created for specific model families: *LVEF Two-Stream TFRecords* for Two-Stream architectures and *LVEF I3D Regression TFRecords* for all other models [27]. Dataset access and management were handled via kagglehub. Training employed EarlyStopping and LearningRateScheduler callbacks.

**Evaluation Protocol:** Model performance was evaluated using root mean squared error (RMSE), mean absolute error (MAE), and the coefficient of determination ($R^2$). To support detailed qualitative analysis, we generated prediction-versus-ground-truth plots, Bland–Altman plots, and training history curves for each model. Generalization performance was assessed by comparing train–test metric gaps and validation–training loss discrepancies. Particular attention was paid to performance and precision within the clinically critical 40–50% EF range.

All trained models were serialized in .keras format, alongside training histories, evaluation plots, and performance metrics. To ensure consistency across models, all metrics were recomputed on the best epoch using the training, validation, and test sets.

**Architectural Variations and Hyperparameter Analysis:** Architectural experiments focused on two primary factors: normalization strategy and convolutional kernel size. All models were evaluated using both Batch Normalization (BatchNorm) and Layer Normalization (LayerNorm). BatchNorm normalizes activations across the batch and spatial dimensions on a per-feature basis, improving training stability but introducing batch-size dependencies. In contrast, LayerNorm normalizes across channels and spatial dimensions on a per-sample basis, making it independent of batch size and better suited for recurrent architectures such as LSTMs [34].

For I3D and selected Fusion models, we additionally compared 1×1×1 and 3×3×3 convolutional kernels in the *conv2* layer of the model stem. Models achieving a test RMSE ≤ 8.5% were designated as *well-performing*. This threshold was chosen to approximate expert-level clinical performance, falling within approximately 1.0 RMSE of reported human variability. The Results section prioritizes these models, while auxiliary experiments are reported in the Appendix.

**I3D Architectures:** Our I3D models were adapted from the LVEF I3D architecture provided in the Omdena Kaggle repository [22], originally designed for categorical EF severity classification using EchoNet-Dynamic videos [21]. Unlike segmentation-based approaches, this model operates directly on echocardiogram clips without semantic annotations. The architecture consists of a stem for early feature extraction, followed by three inception modules (Modules 3–5) and a classification head.

Each inception module follows the four-branch structure of Inception-v1 [15], while incorporating improvements from Inception-v2, including the replacement of 5×5 filters with stacked 3×3 filters and consistent use of BatchNorm and ReLU activations [16]. To accommodate smaller input dimensions, the initial downsampling convolution present in standard I3D architectures was omitted [17].

To repurpose the network for regression, we replaced the classification head with a single linear neuron, optimized using mean squared error (MSE) loss [22,25]. Following an ablation study (Appendix), a Global Average Pooling (GAP) + Dropout + Dense regression head was selected. Two variants were evaluated: I3D-original, retaining all inception modules, and I3D-mini, which includes only the stem and Inception Module 3.

**Two-Stream Models:** Two-Stream models were derived from the corresponding Omdena implementation [21,26]. Unlike the original Two-Stream formulation that relies on optical flow [18], the temporal stream in our implementation uses frame differences, substantially reducing computational cost at the expense of explicit motion magnitude and direction information. To encourage stream specialization, different halves of the input clip were provided to each stream [26].

Rather than fusing class scores, intermediate predictions from each stream were concatenated and passed through a dense layer to generate the final EF estimate [26]. Input dimensions were 28 × 112 × 112 × 1 for the spatial stream and 27 × 112 × 112 × 1 for the temporal stream. Performance was improved through increased learning rates, L1/L2 regularization, and the adoption of LayerNorm.

**Fusion Architectures:** Inspired by the Two-Stream I3D framework of Carreira *et al.* [17], we developed a family of Fusion models combining I3D and Two-Stream components in multiple configurations.

Combination models replaced the standard CNNs in the Two-Stream architecture with I3D-mini modules. To mitigate feature loss from multiple regression heads, we introduced New Combination models, which remove stream-specific heads and instead concatenate feature maps prior to a shared regression head.

We also proposed Dual-Input Fusion models, which process full-resolution spatial and temporal inputs through a shared I3D backbone, followed by task-specific convolutional layers. A Dual-Input Truncated variant concatenates feature maps immediately after the I3D-mini modules, mirroring the New Combination design while sharing early representations.

Finally, Single-Input Fusion models process a single spatial input through an I3D-mini module before splitting the resulting feature map into two streams. These streams are then processed independently and fused for regression. While simpler, this design requires stream specialization to emerge implicitly rather than being architecturally enforced.

**CNN–RNN Architectures:** We adapted a CNN–RNN video classification framework [28], originally trained on the UCF101 dataset [29], for LVEF regression. The architecture applies a TimeDistributed wrapper [30] to a pretrained Inception-v3 CNN [16], extracting frame-level features that are aggregated using a GRU-based RNN [9,14] and mapped to a scalar EF prediction via a linear regression head. Grayscale input frames were triplicated to simulate three-channel RGB input.

In addition to transfer learning (with frozen and fine-tuned weights; see Appendix), we evaluated models trained from scratch, replacing Inception-v3 with a 2D version of I3D-mini. Both GRU and LSTM units were assessed. We further explored normalization strategies, comparing BatchNorm and LayerNorm in the CNN, and introducing LayerNorm in the RNN. These configurations, denoted “Mixed1” in the Results, follow established best practices [37,38].

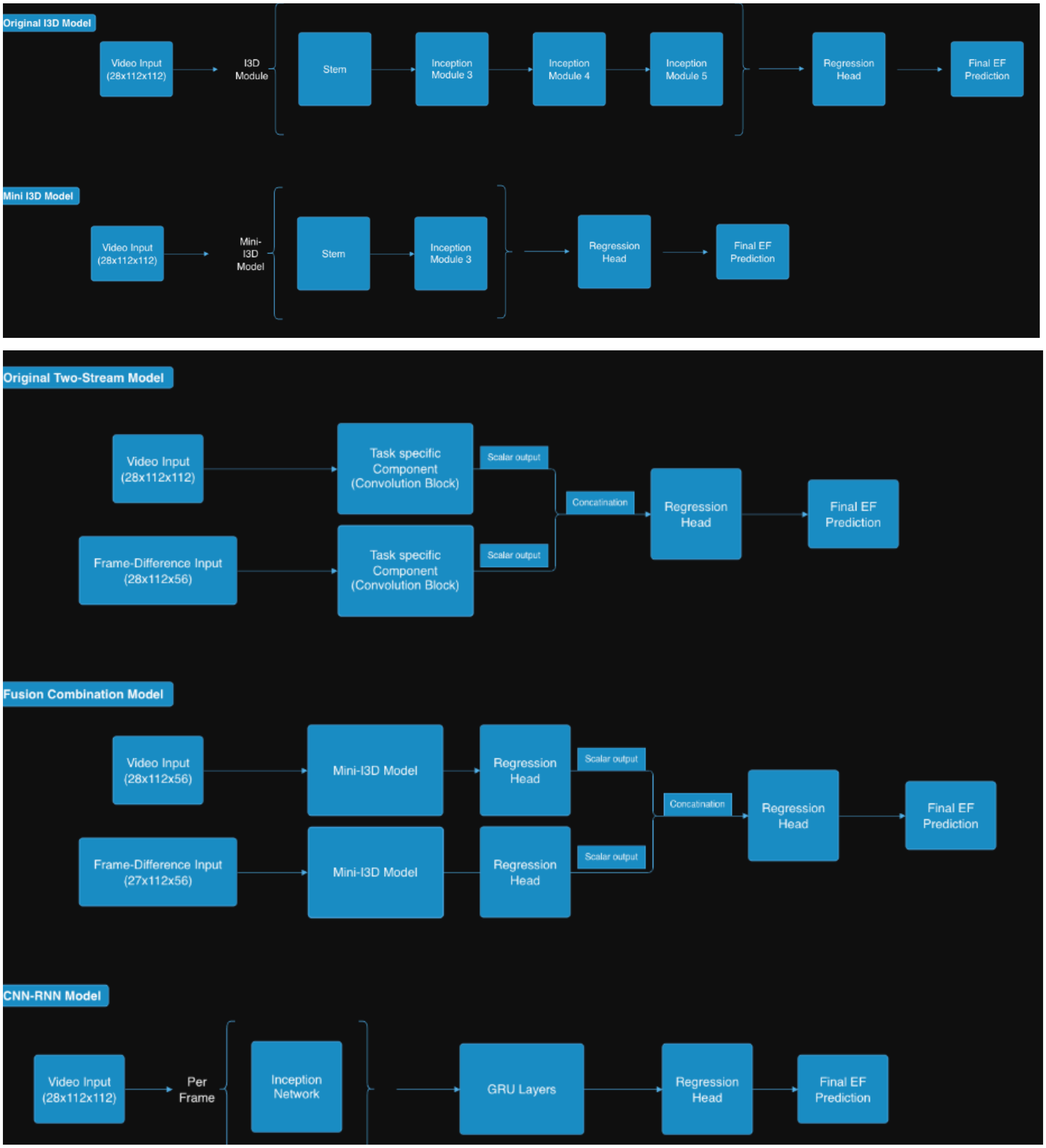

Figure 2 (a): Architectures of I3D, Two-Stream, Fusion Combination and CNN-RNN models used to predict LVEF in this work

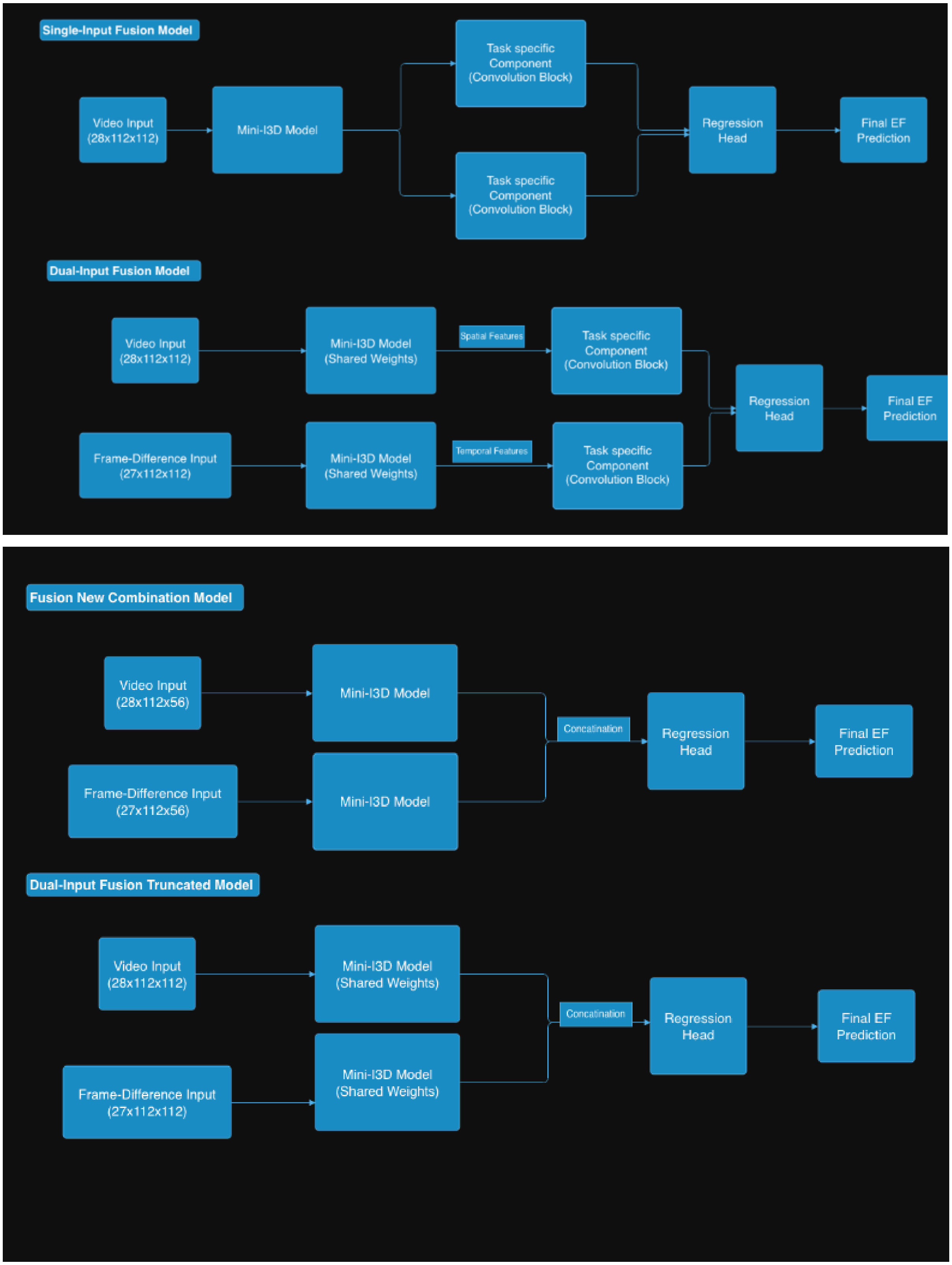

Figure 2 (b): Architectures of different fusion models used to predict LVEF in this work

## Experimental Results

**Experimental Details:** All models were implemented using Google Colab notebooks, leveraging the heart-database dataset and preprocessed TFRecord files available on Kaggle [21,24,27]. Training was conducted for up to 50 epochs, with early stopping applied after 20 epochs without improvement in validation RMSE. Batch sizes were adjusted to accommodate GPU memory constraints and reduce training time: 16 for Two-Stream, Combination, and New-Combination models; 8 for pretrained and CNN–RNN Scratch (LSTM) models; 4 for CNN–RNN Scratch (GRU), Single- and Dual-Input Fusion models, and selected I3D variants; and 2 for the remaining I3D models. An initial learning rate of $1\times10^{-3}$ was used, decayed by a factor of 2 every 10 epochs. To mitigate exploding gradients, the initial learning rate was reduced to $5\times10^{-4}$ for Two-Stream models and $1\times10^{-4}$ for Single- and Dual-Input Fusion models, with gradient clipping applied where necessary [31]. Dropout was applied to all regression heads at a rate of 0.5, except for the Two-Stream (0.05) and CNN–RNN (0.4) models, which retained their original configurations. All models were optimized using the Adam optimizer [32] with mean squared error (MSE) as the loss function. Weight decay and L1/L2 regularization were applied selectively depending on the architecture.

Table 1: Ablation study for I3D Inception modules, includes both I3D-mini and I3D-original configurations

| **Ablation Study for I3D Inception Modules** | **RMSE** | **MAE** | **$R^2$** |
|---|---|---|---|
| Stem | 8.4287 | 6.3985 | 0.5247 |
| Stem + IM3 (I3D-mini) | 7.0906 | 5.3833 | 0.6636 |
| Stem + IM3 + IM4 | 7.4584 | 5.5630 | 0.6278 |
| Stem + IM3 + IM4 + IM5 (I3D-original) | 7.1306 | 5.2933 | 0.6598 |

Table 2: Comparative performance of I3D-mini variants with different final regression heads, all with a linear activation function

| **Changing Regression Heads for I3D-mini** | **RMSE** | **MAE** | **$R^2$** |
|---|---|---|---|
| Dropout + Flatten + Dense (OG) | 7.2014 | 5.3379 | 0.6530 |
| GAP + Dropout + Dense (A) | 7.0906 | 5.3833 | 0.6636 |
| Dropout + 1x1x1 Conv + GAP (B) | 7.1878 | 5.2721 | 0.6543 |
| GAP + Dense + Dropout + Dense (C) | 7.4074 | 5.4792 | 0.6329 |

Table 3: Comparative performance of I3D-mini variants with different Conv2 kernel sizes, all utilizing a GAP+Dropout+Dense regression head

| **Changing Conv2 kernel size** | **RMSE** | **MAE** | **$R^2$** |
|---|---|---|---|
| 1x1x1 (B1) | 7.0906 | 5.3833 | 0.6636 |
| 3x3x3 (B2) | 6.7954 | 5.0103 | 0.691 |
| 3x1x1 (B3) | 6.9925 | 5.1247 | 0.6729 |
| 2 3x3x3 (B4) | 7.0053 | 5.3681 | 0.6717 |

**Performance of I3D Models:** The I3D ablation study examining the contribution of individual Inception Modules (IMs) showed that removing IM4 and IM5 did not result in a significant change in performance, whereas adding IM4 or removing IM3 led to clear performance degradation (see Table 1). These findings indicate that IM3 constitutes the core representational component of the I3D architecture for LVEF estimation, while IM4 and IM5 contribute little additional benefit for this task. Accordingly, both I3D-mini and I3D-original architectures were retained for subsequent experiments. Notably, I3D-mini consistently exhibited smaller gaps between training and test metrics, as well as between training and validation loss, suggesting reduced overfitting compared to I3D-original. Given its superior generalization performance and lower computational and memory requirements, I3D-mini is therefore better suited for LVEF prediction.

The regression head ablation study for I3D-mini demonstrated that the Global Average Pooling (GAP) + Dropout + Dense configuration outperformed the original Dropout + Flatten + Dense head, as well as other evaluated variants (see Table 2). The use of GAP likely promotes more effective spatial aggregation and information retention than flattening, as reflected by a reduced generalization gap. Additional architectural components, such as a 1×1×1 convolution or extra Dense layers, did not lead to further performance gains; however, the inclusion of a single additional Dense layer was found to mitigate overfitting in certain configurations.

Modifying the kernel size of the Conv2 layer, originally configured with a 1×1×1 kernel, yielded substantial performance improvements, with the 3×3×3 configuration producing the largest gains for I3D-mini (see Table 3). Increasing the kernel size enables the convolutional layer to incorporate spatially and temporally neighboring information, thereby expanding the effective receptive field and enhancing feature extraction. The observed performance progression from 1×1×1 (B1) to asymmetric 3×1×1 (B3) and finally to symmetric 3×3×3 (B2) underscores the importance of jointly modeling spatial and temporal context. However, further increases in receptive field size such as stacking additional 3×3×3 convolutions resulted in degraded performance. This likely reflects a mismatch between theoretical and effective receptive fields, wherein increased model complexity reduces the proportion of the input that meaningfully contributes to each activation [38], ultimately leading to overfitting. The 3×3×3 kernel was thus identified as the optimal trade-off between representational capacity and generalization. This modification was adopted across most architectures, with the exception of CNN–RNN models (which employ 2D convolutions) and Two-Stream models (which do not include an I3D stem).

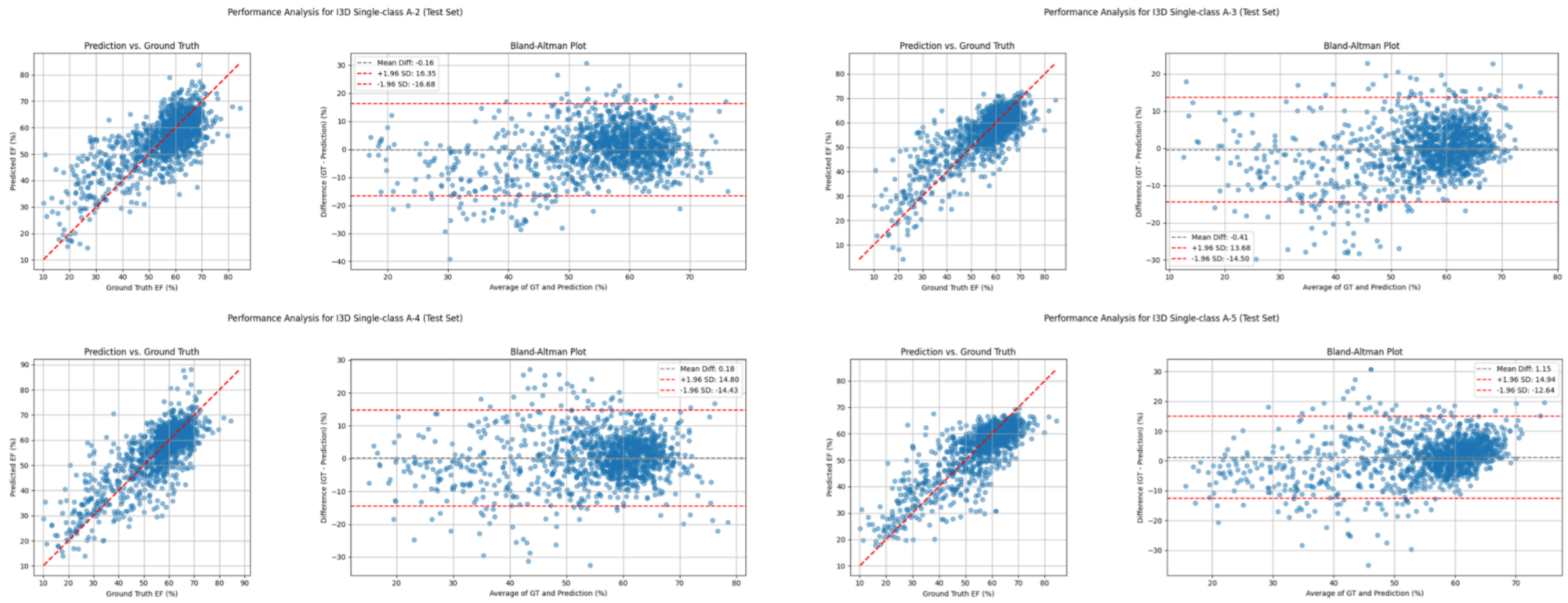

(a)

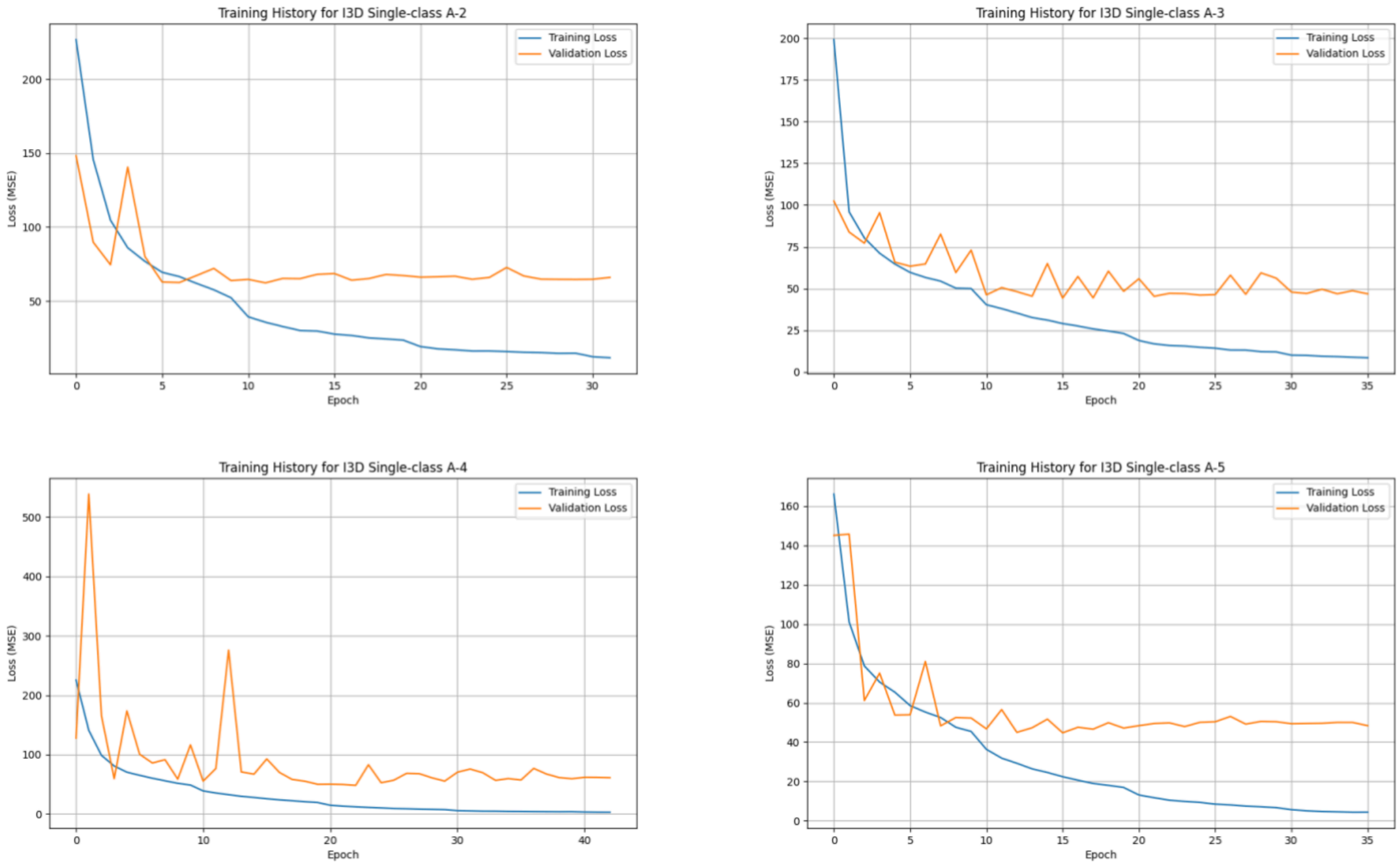

(b)

Figure 3: (a) Performance Analysis and (b) Learning curves for I3D Ablation Study

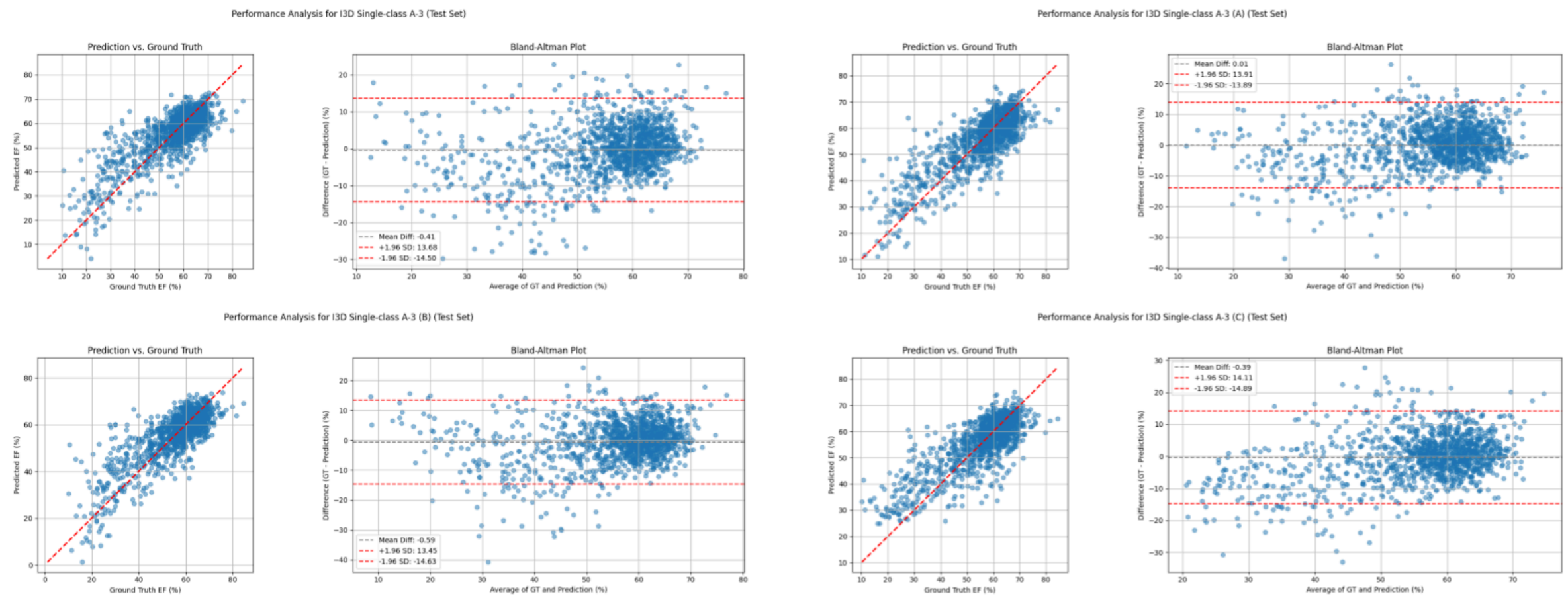

(a)

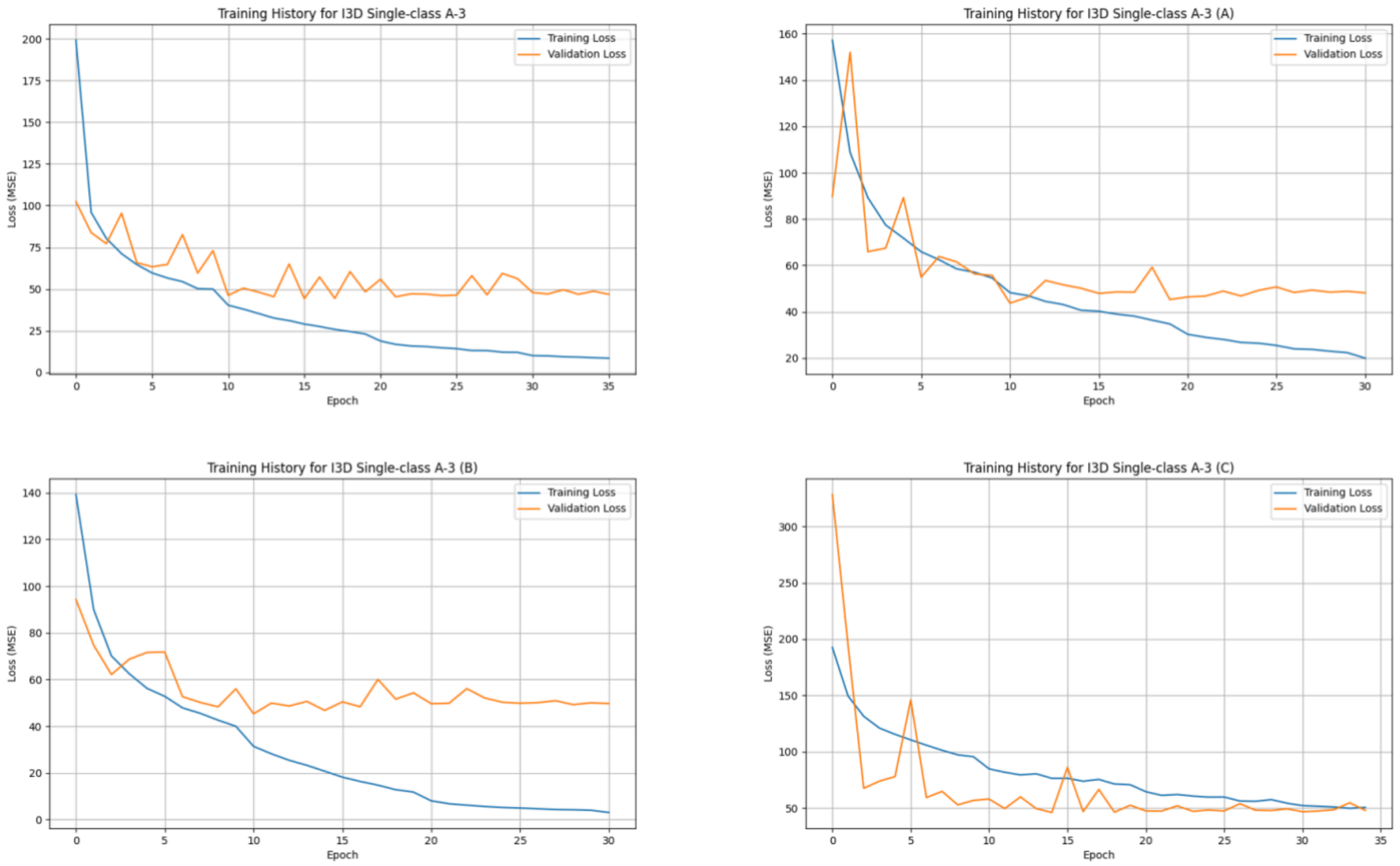

(b)

Figure 4: (a) Performance Analysis and (b) Learning curves for I3D-mini models with various regression heads

Table 4: Performance comparison for I3D-original and I3D-mini variants with various normalization techniques and Conv2 kernel sizes

| **I3D Model Metrics (Test Set)** | **Original** | | | **Mini** | | |
|---|---|---|---|---|---|---|
| | **RMSE** | **MAE** | **$R^2$** | **RMSE** | **MAE** | **$R^2$** |
| BatchNorm (B1) | 7.1306 | 5.2933 | 0.6598 | 7.0906 | 5.3524 | 0.6634 |
| LayerNorm (L1) | 12.230 | 8.9940 | -0.0007 | 7.7140 | 5.7198 | 0.6019 |
| 3x3x3 Conv2 + BatchNorm (B2) | 6.8985 | 5.0972 | 0.6816 | 6.7954 | 5.0103 | 0.6910 |
| 3x3x3 Conv2 + LayerNorm (L2) | 12.2295 | 9.0004 | 0.0007 | 7.060 | 5.2197 | 0.6669 |

Overall, the I3D family of models achieved the strongest performance. The I3D-original model recorded a test RMSE of 7.13, while the I3D-mini model had a test RMSE of 7.09 (see Table 4). Furthermore, performance was highly sensitive to the kernel size of the conv2 layer. Testing various kernel sizes found that the 3x3x3 kernel provided the greatest performance increase, outperforming the original 1x1x1 configuration by 0.25-0.3 RMSE across all I3D models. Conversely, substituting BatchNorm with LayerNorm was found to have a negative effect on performance, with the loss being more pronounced for I3D-original than for I3D-mini. Models such as I3D OG_L1 and I3D OG_L2 suffered from predictive collapse, degenerating into mean-predictors and outputting a single constant value for all samples ($R^2 \approx 0$).

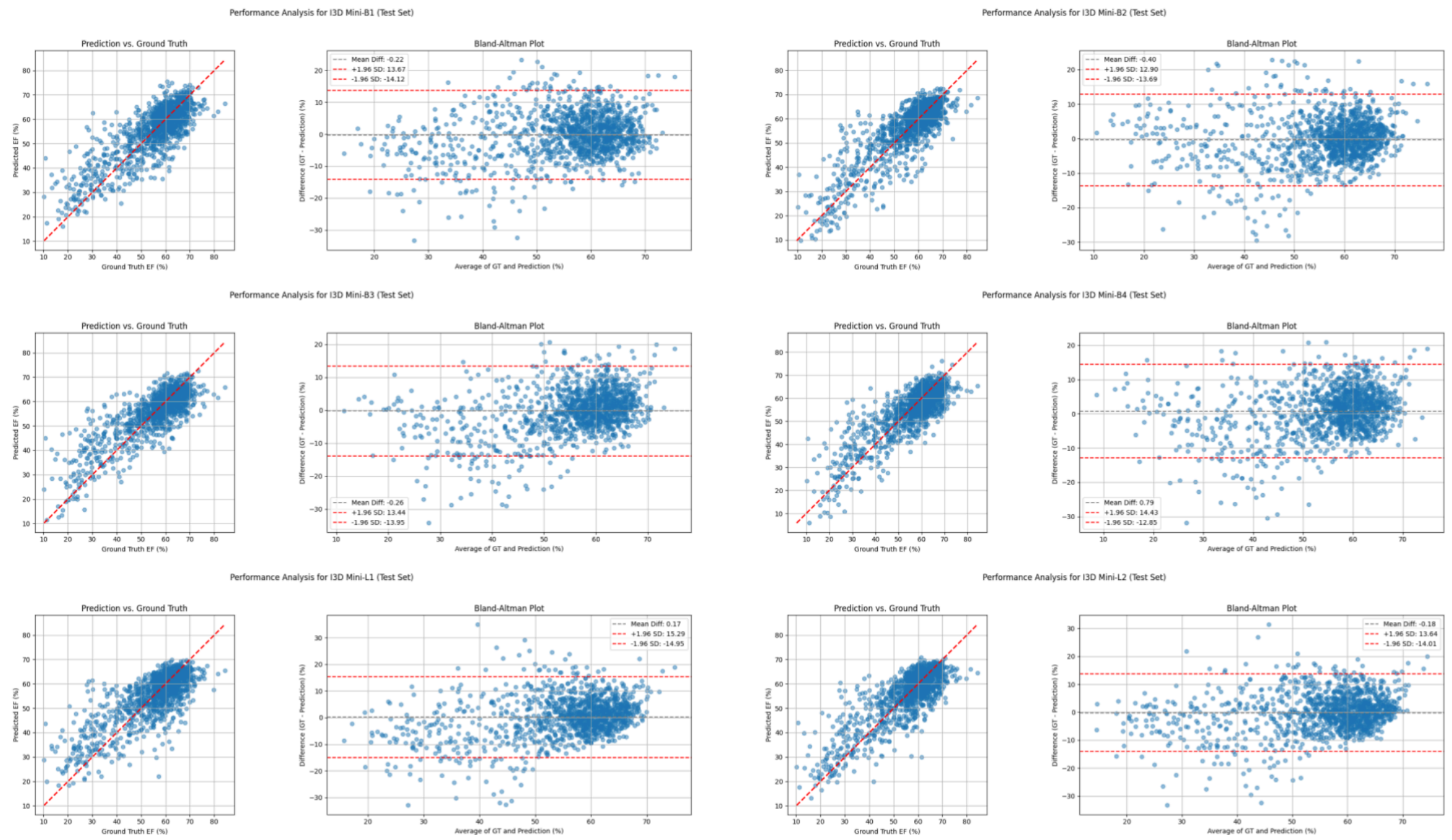

(a)

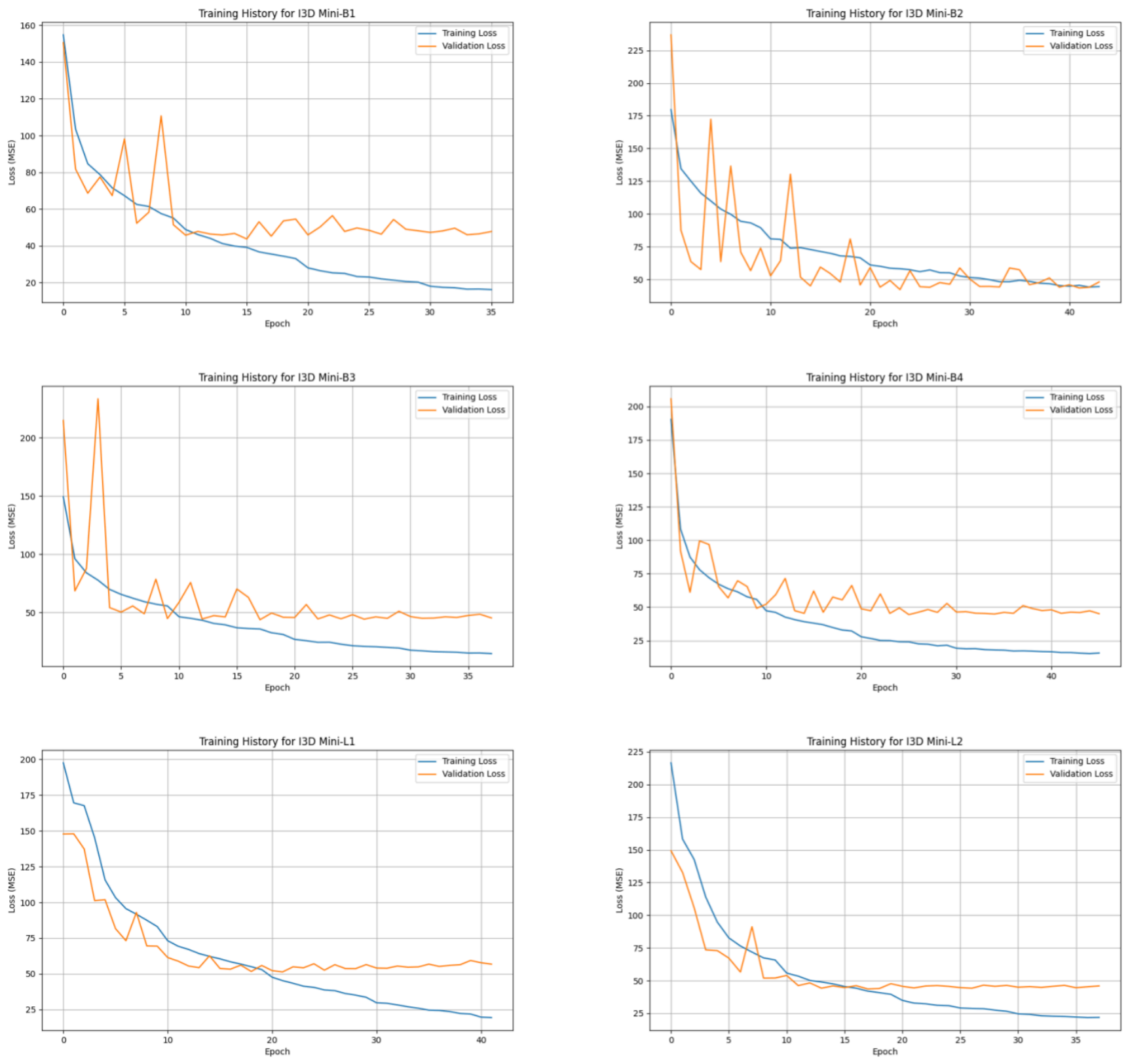

(b)

Figure 5: (a) Performance Analysis and (b) Learning curves for I3D-mini models

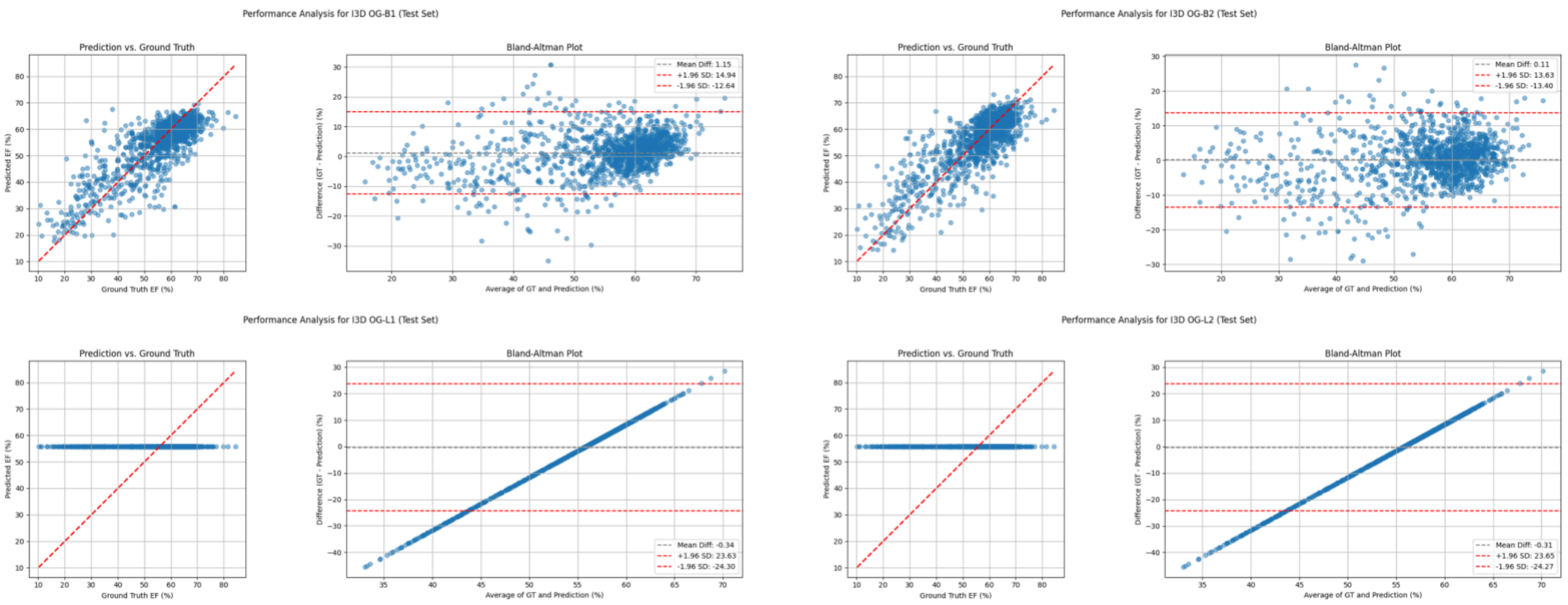

(a)

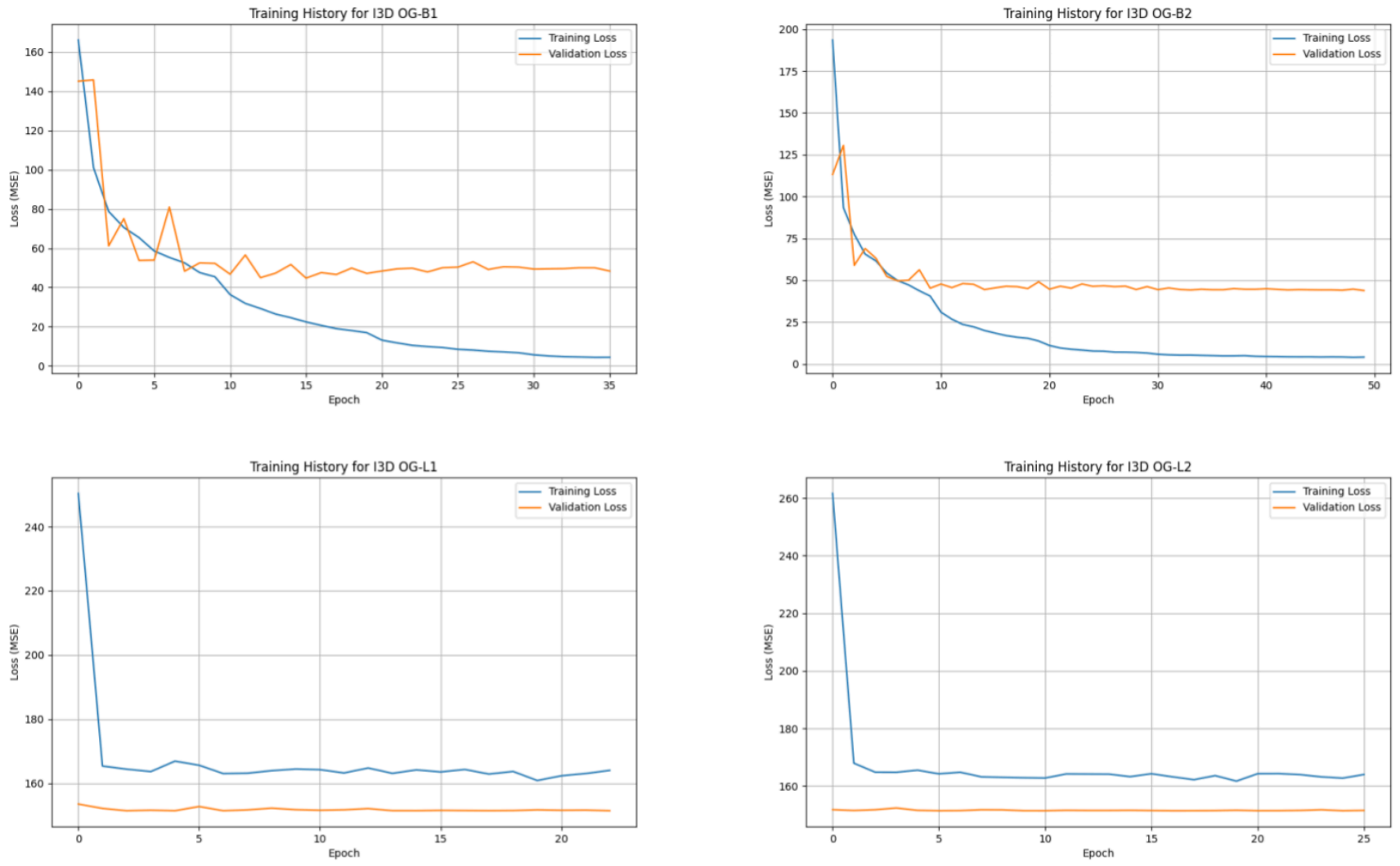

(b)

Figure 6: (a) Performance Analysis and (b) Learning curves for I3D-original models

**Performance of Two Stream Models:** The Two-Stream models demonstrated moderate performance but consistently fell just below the threshold for well-performing architectures, with the best-performing variant (S-1) achieving a test RMSE of 8.85 (see Table 5). In contrast to the I3D-based models, replacing Layer Normalization with Batch Normalization led to a noticeable degradation in performance. Across variants, these models exhibited difficulty in accurately predicting cases within the clinically critical low-to-mid LVEF range (30–50%), showing a pronounced tendency toward regression to the mean and systematic overestimation in this interval. This behavior directly contributed to their elevated error metrics. Overall, these findings indicate that the Two-Stream architecture, as evaluated here, lacks the representational capacity and robustness required for precise LVEF estimation, rendering it less suitable for this task.

Table 5: Performance metrics for Two-Stream variants with various normalization techniques

| Two-Stream Model Metrics (Test Set) | Original | | |
|---|---|---|---|
| | RMSE | MAE | $R^2$ |
| BatchNorm (B1) | 9.7161 | 7.3460 | 0.3697 |
| LayerNorm (L1) | 8.8489 | 6.6387 | 0.4758 |

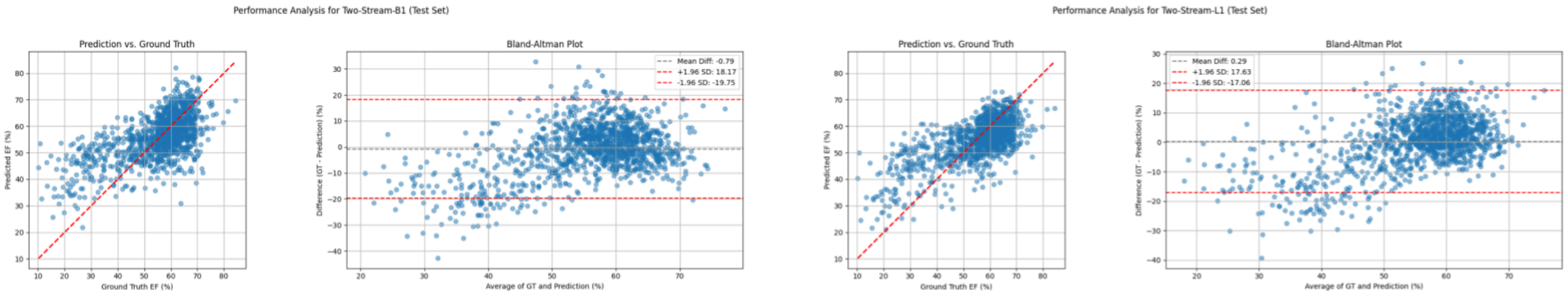

(a)

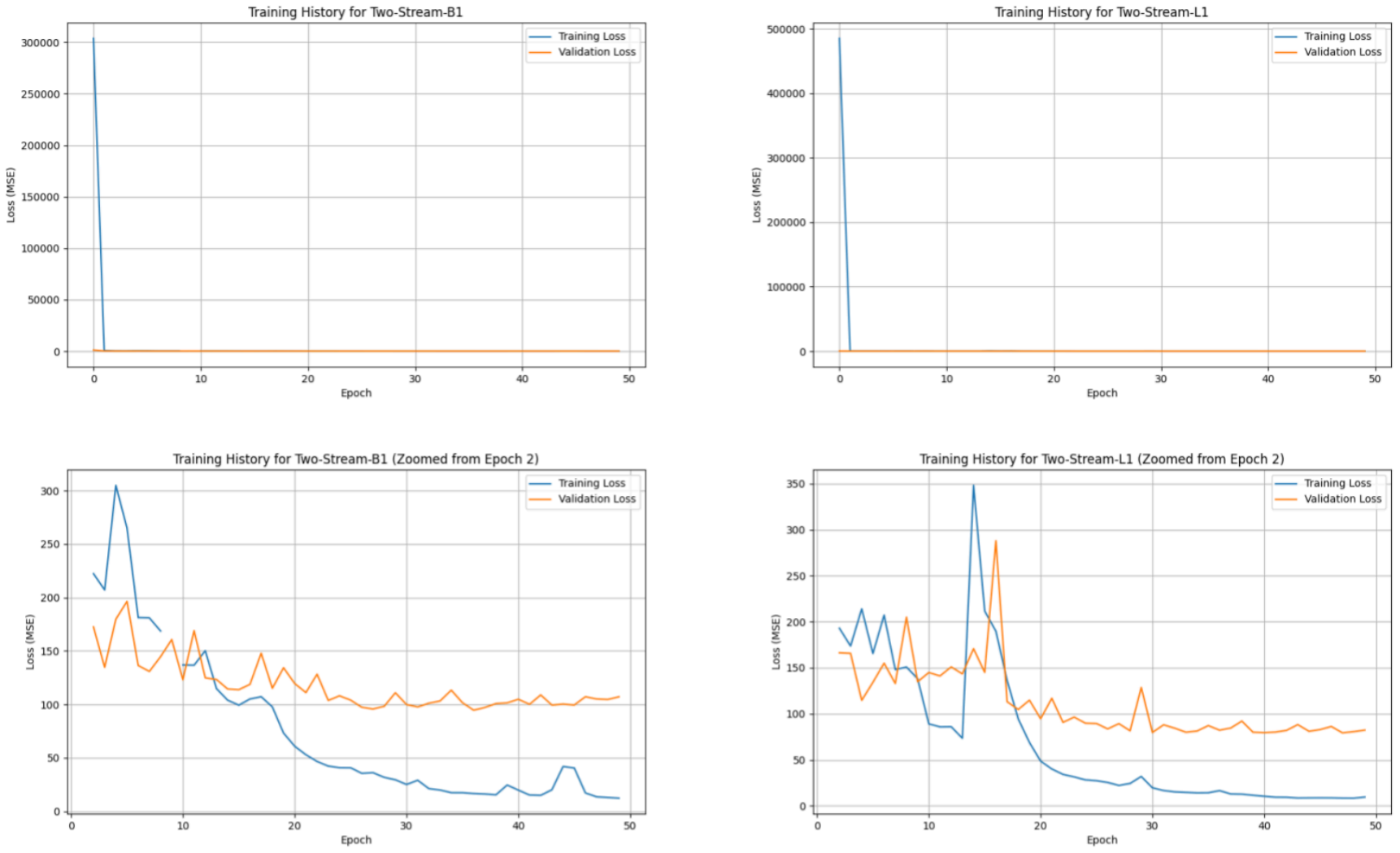

(b)

Figure 7: (a) Performance Analysis and (b) Learning curves for Two-Stream models

**Performance of Fusion Models:** The Fusion models exhibited heterogeneous performance, revealing several notable trends. The initial Combination architecture performed poorly, achieving a test RMSE of 9.14 when using Layer Normalization. This limited performance is likely attributable to the use of half-resolution tensors in each stream and the presence of multiple independent regression heads, which may have resulted in feature fragmentation and information loss.

Replacing the multiple heads with a single shared regression head in the New Combination model led to a substantial improvement, yielding a test RMSE of 7.96 when combined with a 3×3×3 Conv2 kernel and Layer Normalization. More pronounced gains were observed with the Single-Input and Dual-Input Fusion architectures, which achieved test RMSE values of 7.02 and 7.64, respectively, both when paired with Batch Normalization. These results suggest that stronger feature sharing and deeper joint representation learning significantly benefit LVEF estimation.

In contrast, the Dual-Input Truncated variants performed markedly worse, with a test RMSE of 11.45 (see Tables 6–7). This degradation likely stems from premature truncation of the shared backbone, which limits the model's capacity to learn sufficiently rich spatiotemporal representations. Overall, these findings highlight the importance of both architectural depth and appropriate fusion strategies when integrating spatial and temporal information for accurate LVEF prediction.

Table 6: Performance comparison of different Fusion variants with various normalization techniques and Conv2 kernel sizes.

| Fusion Model Metrics | New Combination | | | Dual Input | | | Single Input | | |
|---|---|---|---|---|---|---|---|---|---|
| | RMSE | MAE | $R^2$ | RMSE | MAE | $R^2$ | RMSE | MAE | $R^2$ |
| BatchNorm (B1) | 8.0861 | 6.1683 | 0.5625 | 7.6384 | 6.3294 | 0.5016 | 7.0173 | 5.2039 | 0.6705 |
| LayerNorm (L1) | 8.0632 | 6.0326 | 0.5620 | 8.7145 | 5.8045 | 0.6096 | 7.4718 | 5.5195 | 0.6265 |
| 3x3x3 Conv2 + BatchNorm (B2) | 8.0909 | 6.0141 | 0.5650 | 7.8777 | 5.9397 | 0.5847 | 7.0846 | 5.2166 | 0.6642 |
| 3x3x3 Conv2 + LayerNorm (L2) | 7.9595 | 5.8767 | 0.5761 | 7.9344 | 5.8794 | 0.5788 | 7.1433 | 5.3555 | 0.6586 |

Table 7: Performance comparison of different Fusion variants with two normalization techniques and Conv2 kernel sizes.

| Fusion Model Metrics | Combination | | | Dual Input Truncated | | |
|---|---|---|---|---|---|---|
| | RMSE | MAE | $R^2$ | RMSE | MAE | $R^2$ |
| BatchNorm (B1) | 10.5860 | 7.9544 | 0.2518 | 11.4526 | 9.2460 | 0.1224 |
| LayerNorm (L1) | 9.1491 | 6.7385 | 0.4400 | - | - | - |

The Single-Input model, despite its simpler architecture, proved to be the most effective Fusion mode. It was the only model in its family capable of accurately regressing low EF values (Figure 9a), and achieved the lowest RMSE among all models retaining the default 1x1x1 Conv2 kernel. This success reinforces the effectiveness of emergent stream specialization, allowing the I3D-mini module to process features naturally rather than relying on multiple inputs, which may hinder progress during training or concatenation. Furthermore, these results indicate that intermediate, task-specific convolutional layers between the I3D-mini module and regression head are a positive architectural addition. Notably, increasing Conv2 kernel size and switching to LayerNorm only improved New Combination performance, with the Dual Input and Single Input models performing optimally with no additional modifications. This implies that downstream convolutional layers may be negatively affected by the increased feature complexity generated by the larger kernel size. Additionally, the poor performance of Dual Input Truncated indicates that a shared I3D backbone is effective only when paired with additional task-specific convolutions.

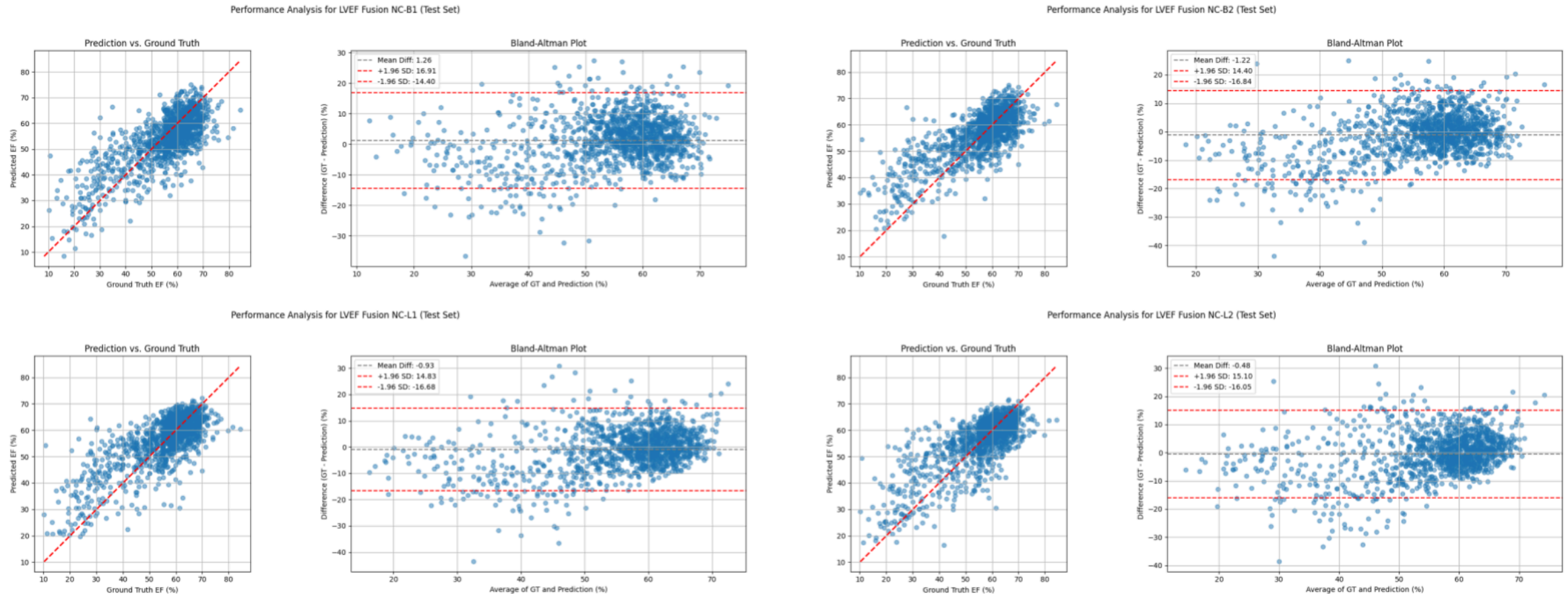

(a)

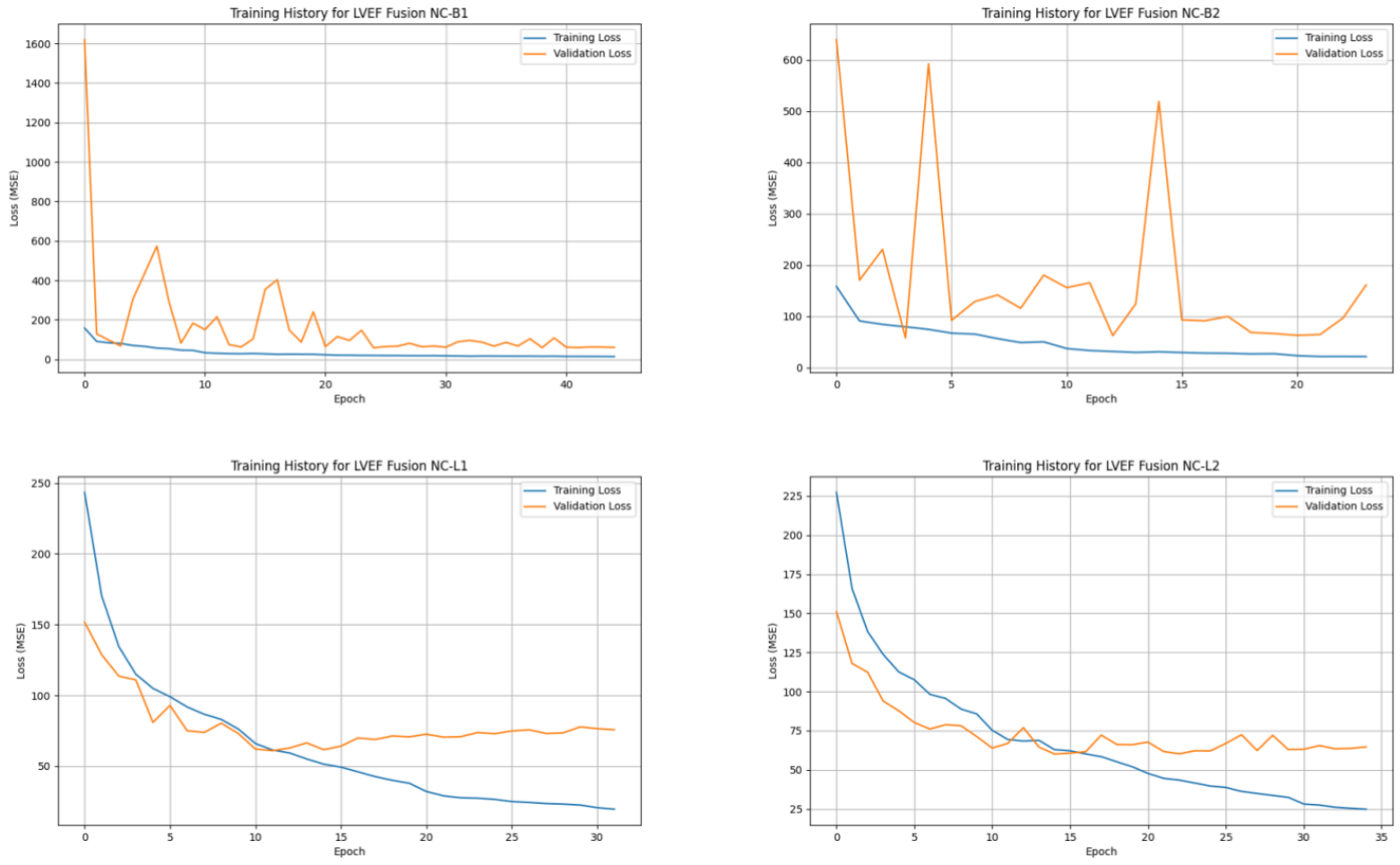

(b)

Figure 8: (a) Performance Analysis and (b) Learning curves for Fusion New combination models

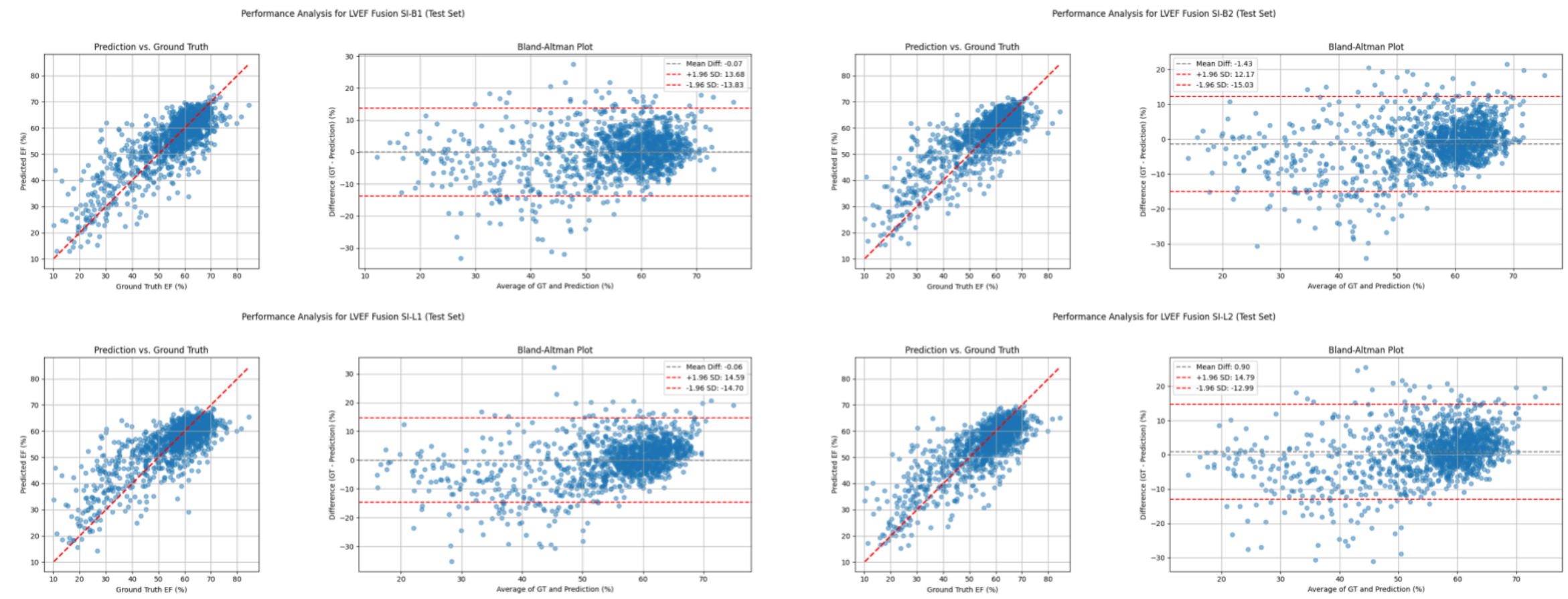

(a)

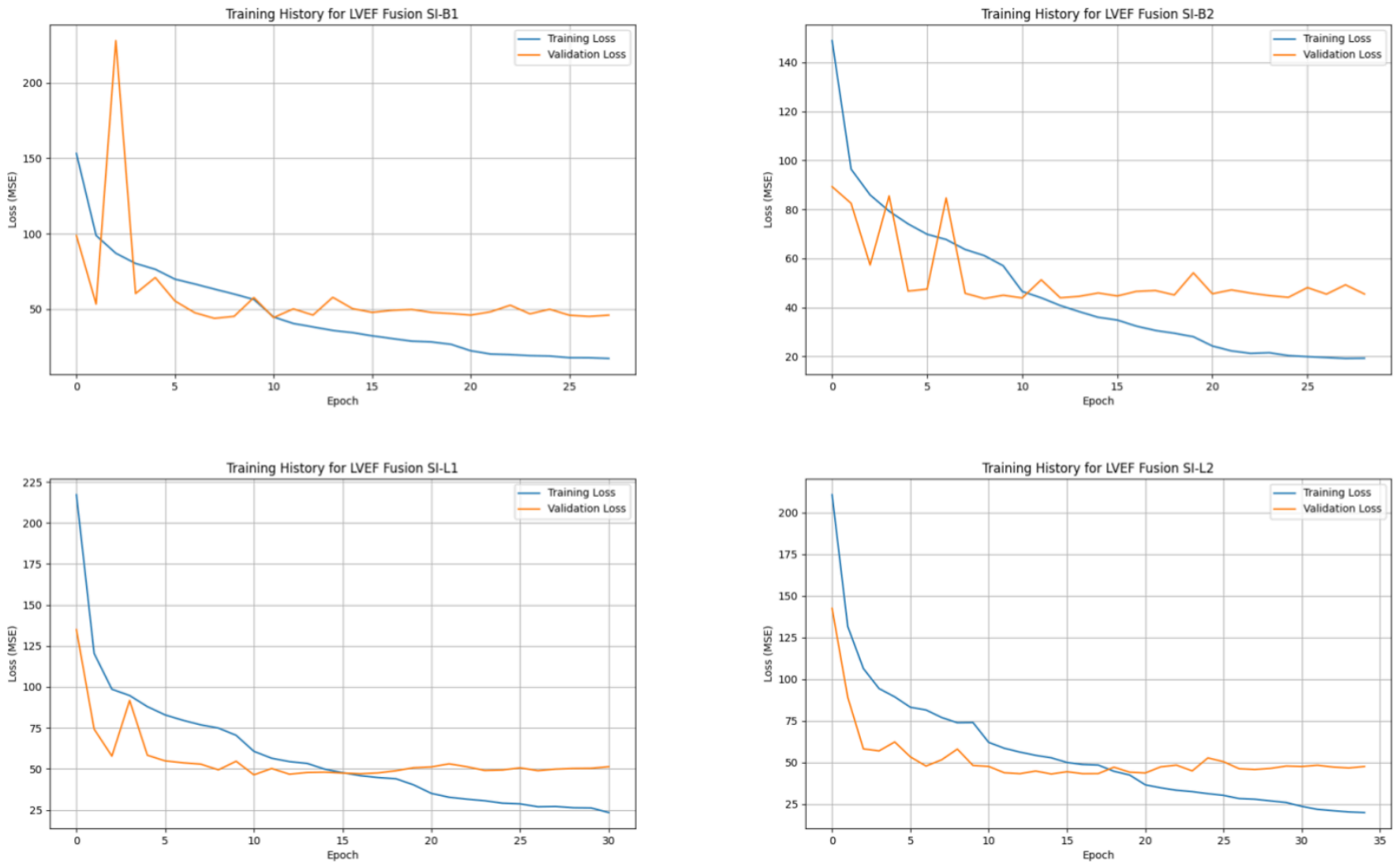

(b)

Figure 9: (a) Performance Analysis and (b) Learning curves for Fusion Single-input models

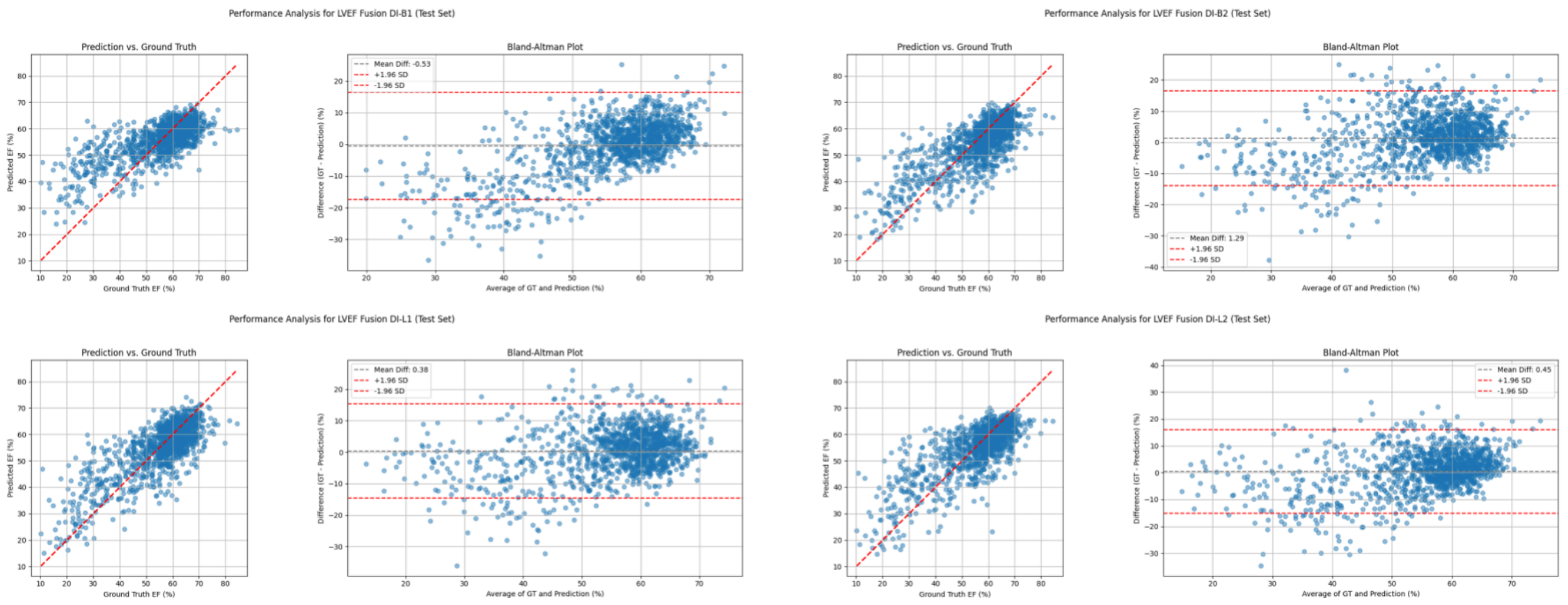

(a)

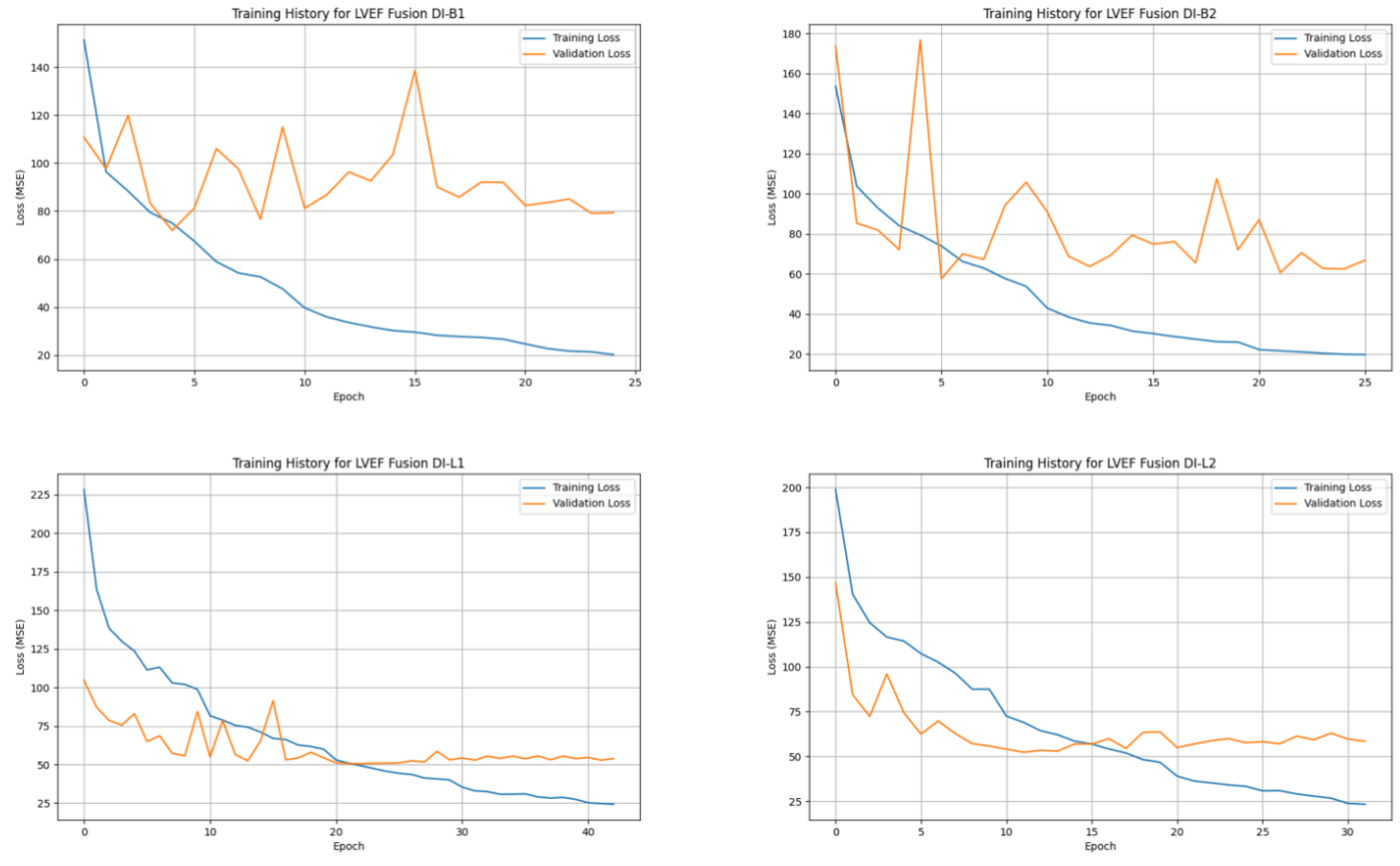

(b)

Figure 10: (a) Performance Analysis and (b) Learning curves for Fusion Dual-input models

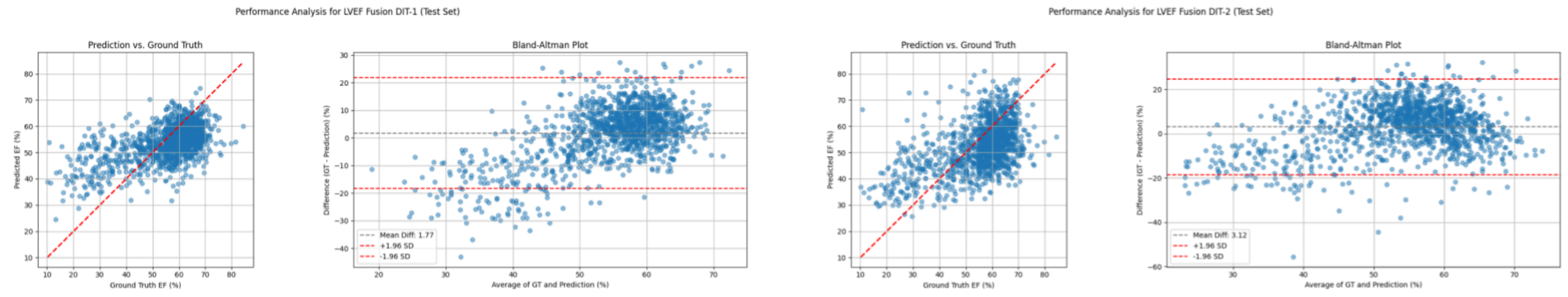

(a)

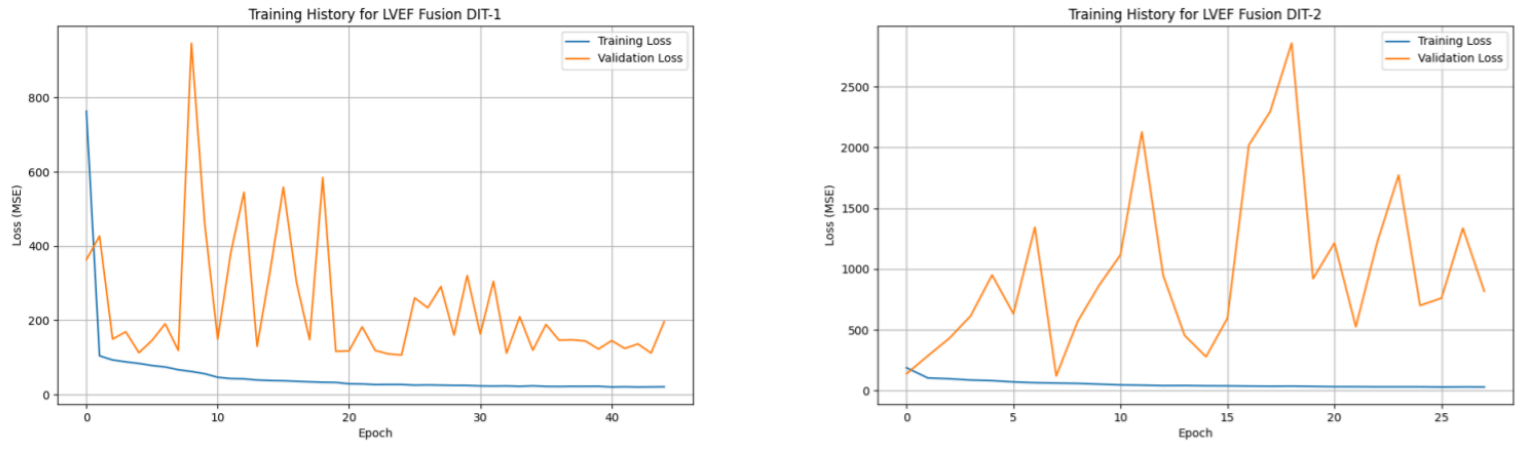

(b)

Figure 11: (a) Performance Analysis and (b) Learning curves for Fusion Dual-input truncated models

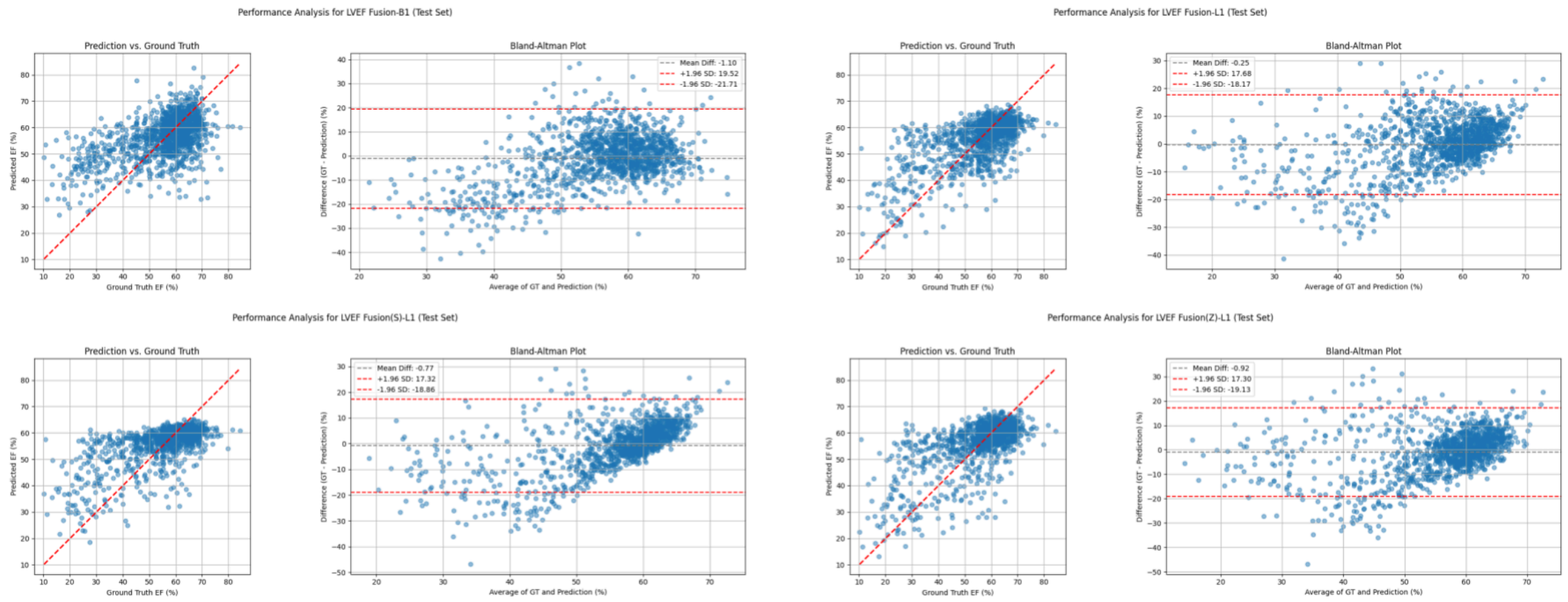

(a)

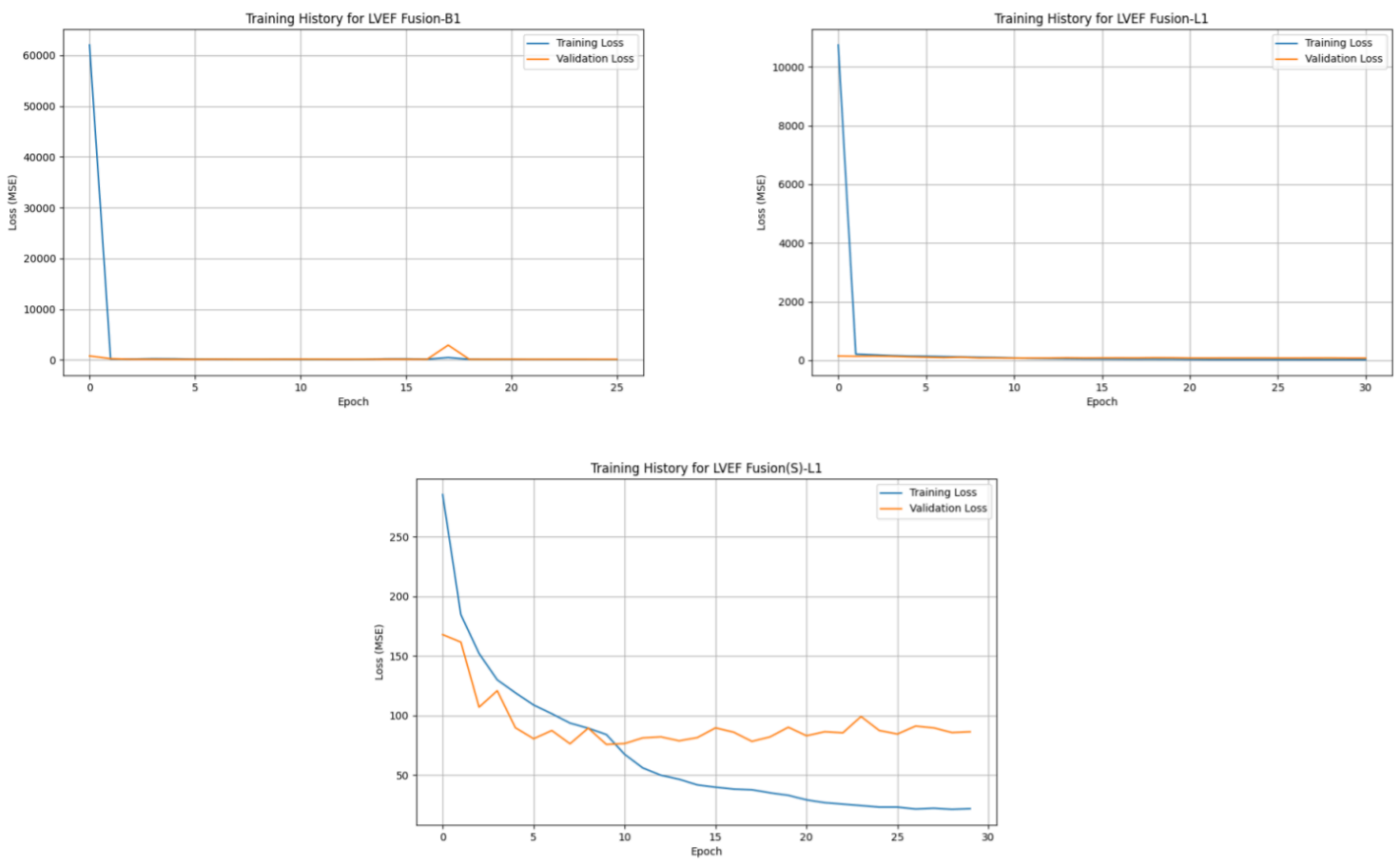

(b)

Figure 12: (a) Performance Analysis and (b) Learning curves for Fusion Combination models

**Performance of CNN-RNN Models:** The CNN–RNN architectures employing a pretrained CNN backbone demonstrated limited effectiveness, achieving a test RMSE of 10.19 with frozen weights and 9.92 after fine-tuning (see Appendix 2 and Table 8). Replacing the pretrained CNN with a 2D variant of I3D-mini resulted in a notable performance improvement, reducing the test RMSE to 8.35. Further gains were observed when GRU layers were replaced with LSTM units, yielding a lowest RMSE of 7.86 (see Table 9). Consistent with trends observed in I3D-based models, Batch Normalization outperformed Layer Normalization in the CNN component. In contrast, the MixedNorm configuration proved ineffective, producing a test RMSE of 10.26 in the LSTM-based variant.

Qualitative inspection of model predictions reveals a persistent lack of granular regression precision (Figure 14a). Although these models can broadly distinguish between low and high LVEF values, they exhibit substantial error in the clinically important intermediate range (30–50%), fail to predict values outside a relatively narrow output band, and do not establish a smooth, continuous relationship between predicted and ground-truth values. While LSTM-based models show marginal improvements, producing more continuous predictions across a wider range, these systemic limitations remain. Consequently, despite achieving reasonable aggregate error metrics, this class of models struggles with precise value regression, significantly limiting its clinical applicability.

Table 8: Performance comparison for CNN-RNN models

| CNN-RNN Pretrained Model Metrics | GRU | | |
|---|---|---|---|
| | RMSE | MAE | $R^2$ |
| CNN_RNN-1 | 10.1854 | 7.4852 | 0.3059 |
| CNN_RNN T(O)-a | 10.4468 | 7.5487 | 0.2698 |
| CNN_RNN T(O)-b(5) | 9.9158 | 7.1689 | 0.3422 |
| CNN_RNN T(O)-b(7) | 10.5819 | 7.5635 | 0.2508 |

Table 9: Ablation study involving different normalization techniques for CNN-RNN scratch models

| CNN-RNN Scratch Model Metrics | GRU | | | LSTM | | |
|---|---|---|---|---|---|---|
| | RMSE | MAE | $R^2$ | RMSE | MAE | $R^2$ |
| BatchNorm (B1) | 8.3539 | 6.4731 | 0.5331 | 7.8644 | 5.7785 | 0.5862 |
| LayerNorm (L1) | 12.2297 | 8.9977 | -0.0007 | 9.4743 | 6.9973 | 0.3994 |
| MixedNorm (Mixed1) | 12.6746 | 10.4873 | -0.0748 | 10.2638 | 8.8397 | 0.2952 |

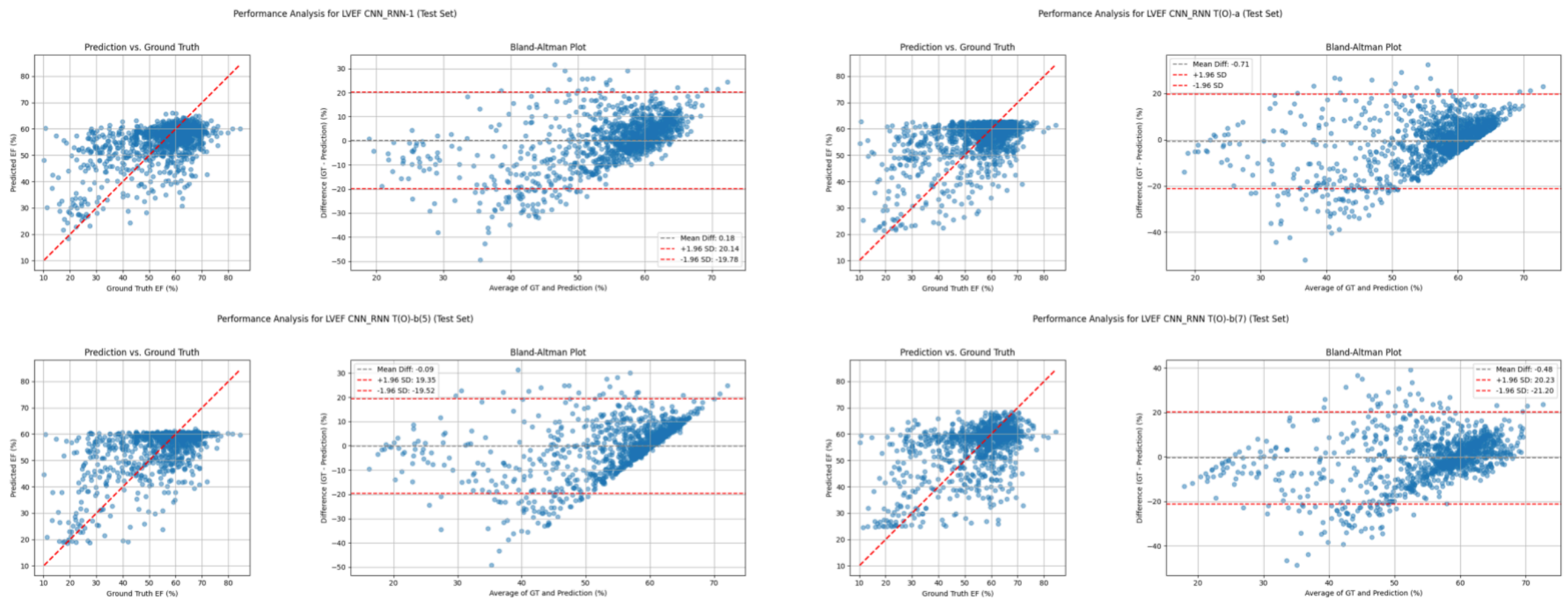

Figure 13: Performance Analysis plots of CNN-RNN models

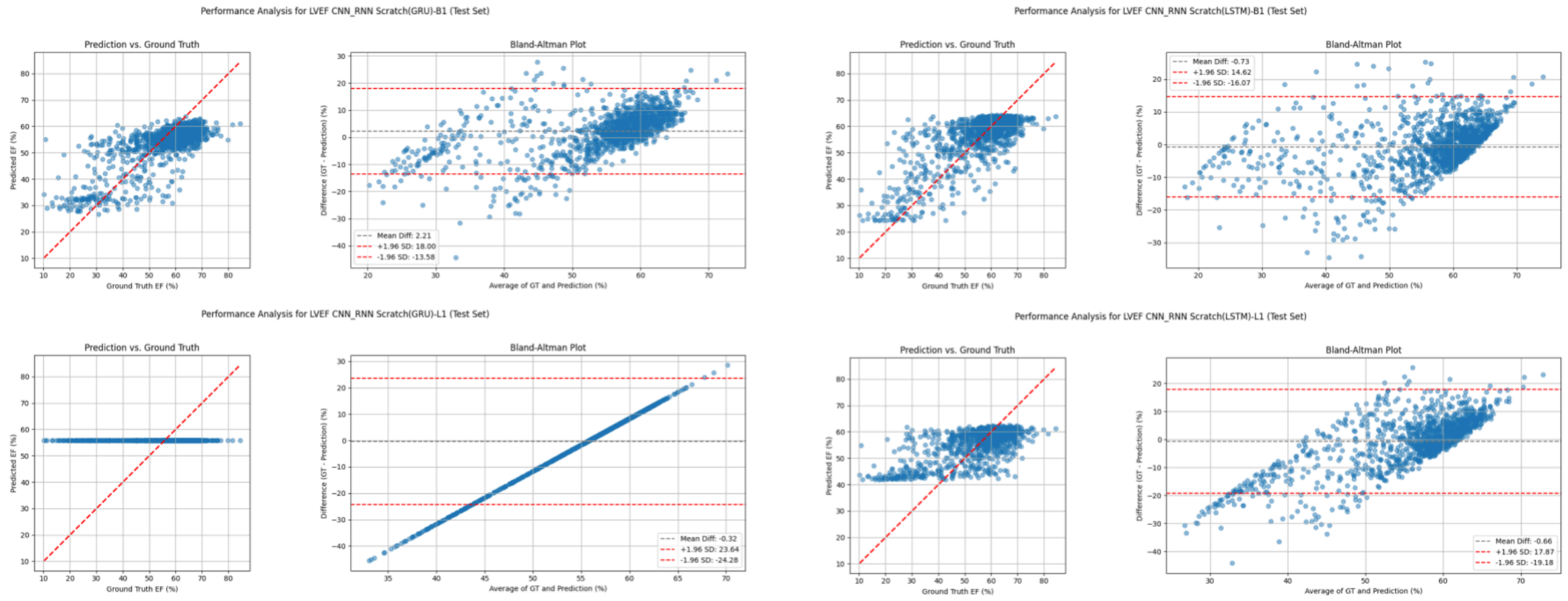

(a)

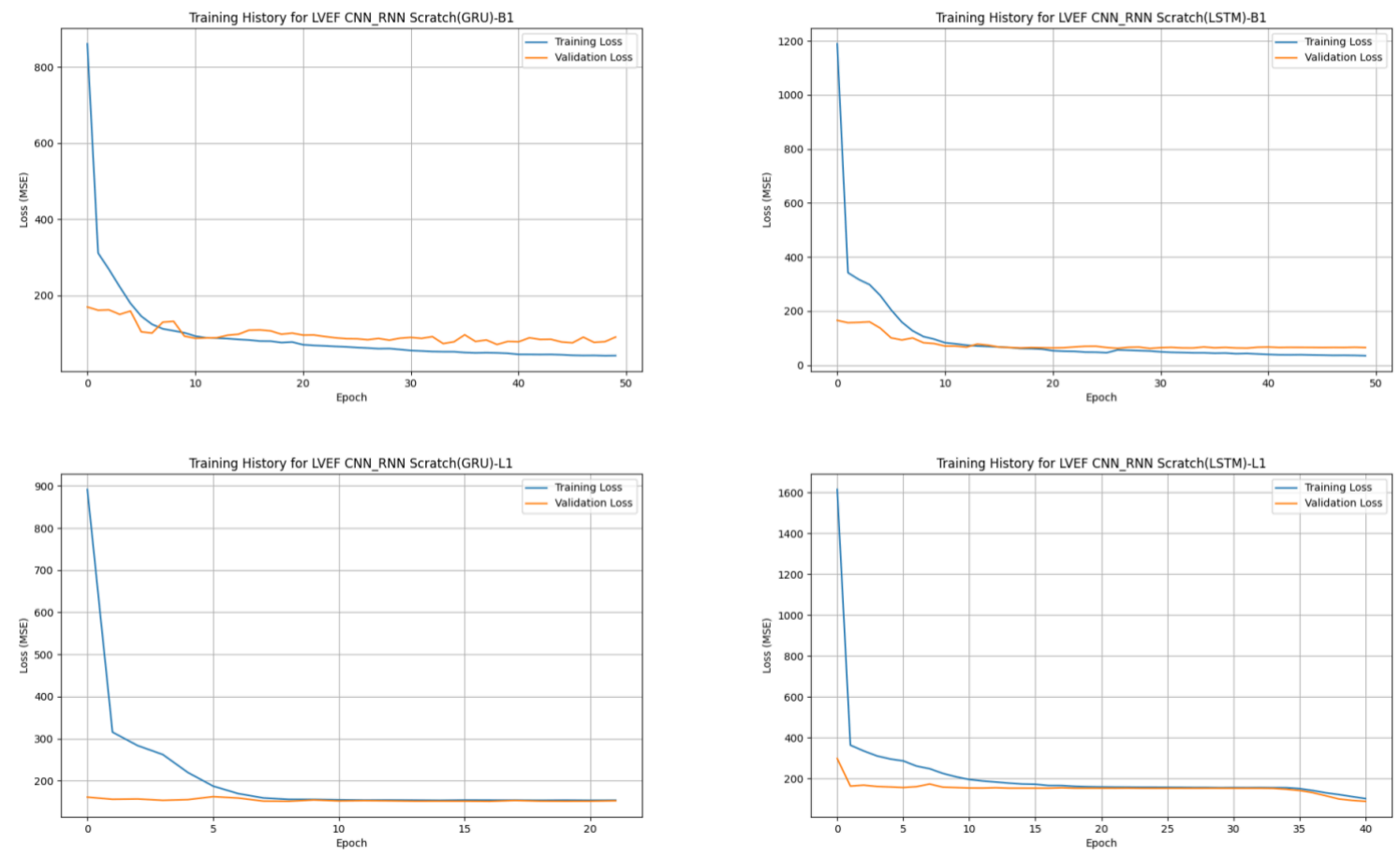

(b)

Figure 14: (a) Performance Analysis and (b) Learning curves for CNN-RNN scratch models

Table 10: Summary of root mean squared error (RMSE) values for the key benchmark models evaluated in this study, comparing the effects of different normalization strategies and Conv2 kernel sizes (for I3D and Fusion architectures).

| Overall performance between models (RMSE) | I3D | | TS | Fusion | | | CNN-RNN | |
|---|---|---|---|---|---|---|---|---|
| | Original | Mini | SC | NC | DI | SI | GRU | LSTM |
| BatchNorm | 7.1306 | 7.0906 | 9.7161 | 8.0861 | 7.6384 | 7.0173 | 8.3539 | 7.8644 |
| LayerNorm | 12.230 | 7.714 | 8.8489 | 8.0632 | 8.7145 | 7.4718 | 12.2297 | 9.4743 |
| 3x3x3 Conv2 + BatchNorm | 6.8985 | 6.7954 | - | 8.0909 | 7.8777 | 7.0846 | - | - |
| 3x3x3 Conv2 + LayerNorm | 12.230 | 7.060 | - | 7.9595 | 7.9344 | 7.1433 | - | - |

Our findings regarding normalization strategies reveal nuanced patterns that in some cases diverge from established observations in the literature [34–36]. Although Batch Normalization (BatchNorm) is often reported to degrade performance and introduce training–inference discrepancies when applied with small batch sizes [35,36], this limitation appears to be mitigated in Inception-based architectures. Owing to their large spatial feature maps, these models provide a sufficiently large number of activation samples per mini-batch, enabling stable estimation of batch statistics despite small batch sizes [37].

Contrary to common expectations that Layer Normalization (LayerNorm) underperforms in convolutional networks [34,35], LayerNorm consistently outperformed BatchNorm in the shallower Two-Stream architectures. We hypothesize that LayerNorm's per-sample normalization is better suited to handling the substantial variability in pixel intensity arising from heterogeneous echocardiography acquisition conditions, which constitutes the dominant source of error in these relatively shallow models.

In contrast, deeper Inception-based models, which concatenate feature maps across multiple branches, benefit more from BatchNorm. In this setting, BatchNorm promotes consistent feature scaling across branches and provides valuable implicit regularization, helping to mitigate overfitting—a key challenge in high-capacity architectures [35,37].

# Discussion

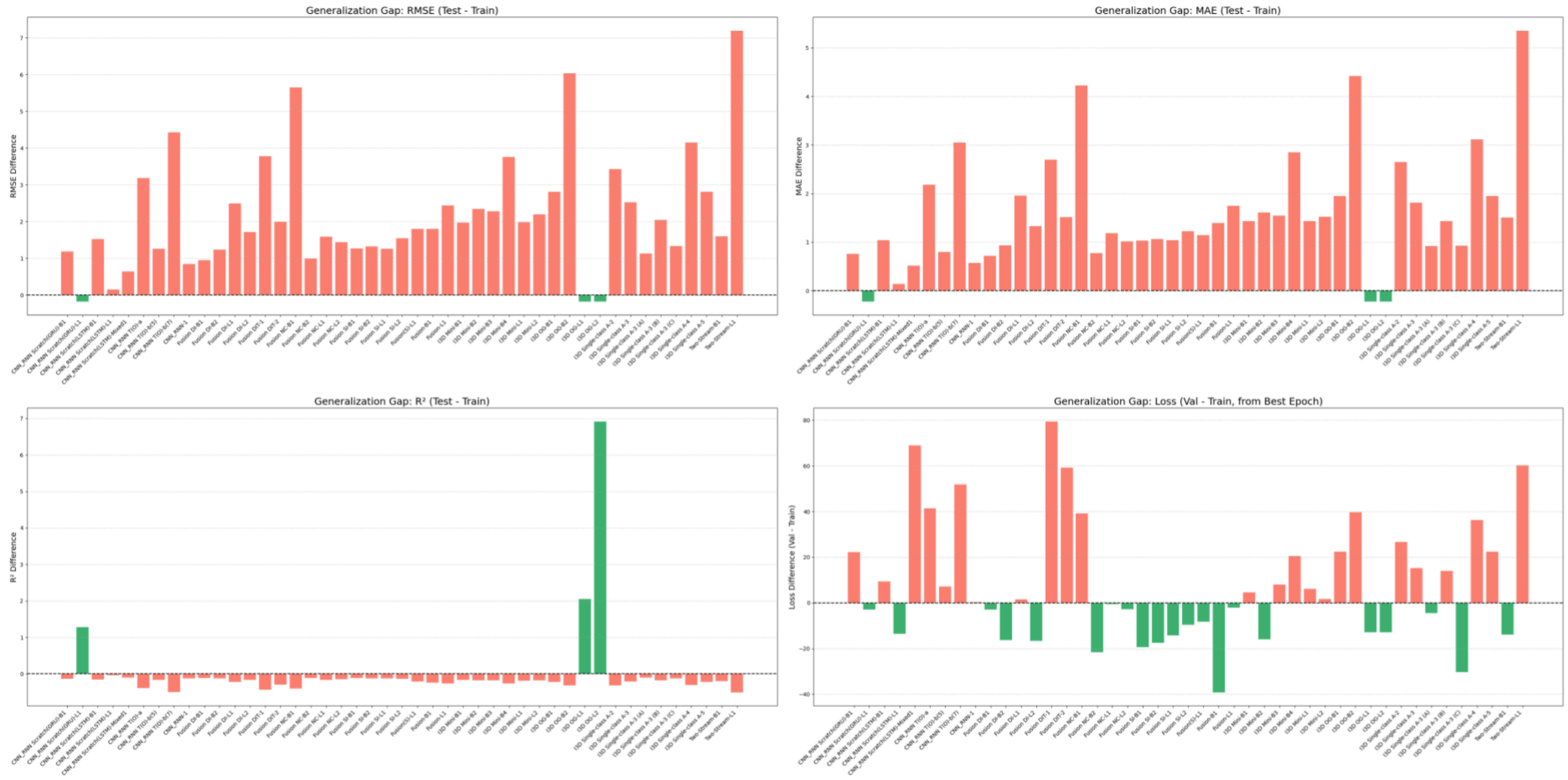

Figure 15: Generalization gap analysis across models. Bars represent differences between training and test metrics (RMSE, MAE, $R^2$) and between training and validation loss. Red bars indicate overfitting, where performance is better on the training set than on the test or validation set, while green bars indicate better performance on the test or validation set than on the training set.

Across all experiments, the models exhibited substantial overfitting (see Fig. 15). While training loss consistently followed a characteristic exponential decay, validation loss was often unstable and tended to plateau early. Even among the best-performing models, predictive accuracy deteriorated in the low LVEF range (10–40%), a clinically critical interval. This degradation is likely attributable to the relative scarcity of samples in this range compared to the more densely represented mid-to-high LVEF range (50–70%). A similar decline in performance was observed for very high LVEF values (>70%). Overall, the models demonstrated a systematic tendency to regress toward the mean, a behavior that was particularly pronounced in Two-Stream and CNN–RNN architectures.

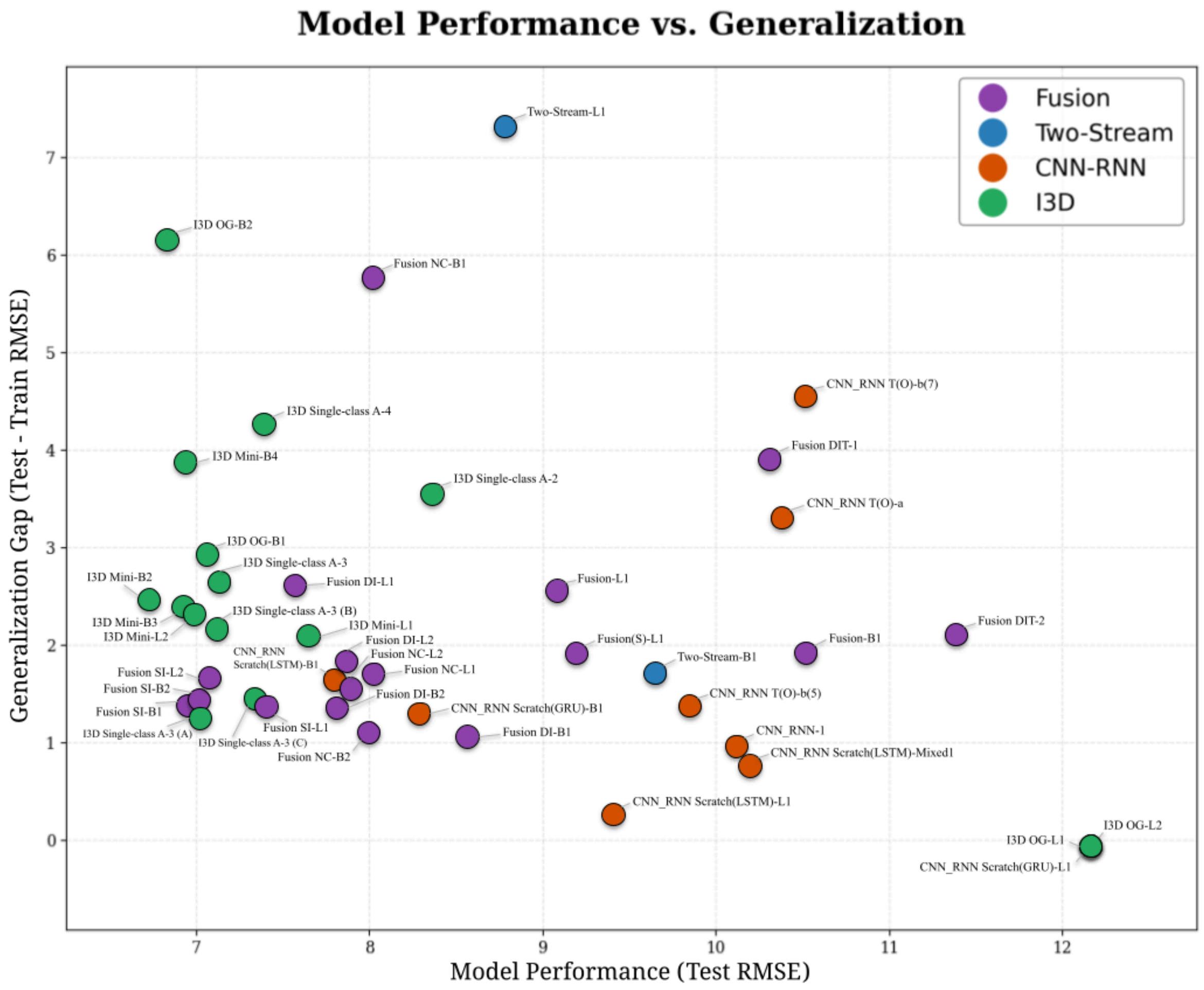

Figure 16: Scatter plot showing Model performance vs Generalization

The I3D family of models consistently achieved the strongest overall performance across all experiments. In particular, I3D-mini demonstrated superior generalization, as evidenced by a narrower divergence between training and test metrics compared with I3D-original. While the Fusion models exhibited heterogeneous performance, the best-performing variants similarly showed reduced generalization gaps. The CNN–RNN Scratch models both outperformed and generalized better than their pretrained counterparts, suggesting that feature representations learned directly from echocardiography data are more effective than those transferred from generic video datasets.

In contrast, the Two-Stream-L1 model, one of the original benchmarks, exhibited the largest generalization gap (approximately 7.2 RMSE), indicating severe overfitting. This behavior suggests that the relatively shallow architecture lacked the capacity to learn robust and discriminative features, instead memorizing training samples. Models clustered in the bottom-right region of the generalization plots serve as a baseline for model collapse; their small generalization gaps are artifacts of their inability to capture meaningful structure in the data rather than indicators of good generalization. Taken together, and excluding clear outliers, these results suggest that a minimum level of architectural depth is necessary to capture relevant spatiotemporal features, whereas excessive complexity exacerbates overfitting.

Accordingly, we hypothesize that the optimal architecture for LVEF estimation must maximize informative feature extraction while minimizing unnecessary complexity. The strong performance of EchoNet-Dynamic, despite its high architectural complexity, highlights the importance of additional inductive biases. In particular, its semantic segmentation task likely guides the model toward clinically relevant anatomical structures, while the use of residual connections mitigates many of the optimization challenges observed in our deeper custom architectures. Future work may benefit from incorporating factorized convolutions to reduce computational burden while preserving expressive capacity. Moreover, techniques designed to alleviate vanishing gradients such as residual connections [41], improved activation functions (e.g., Leaky ReLU), and principled weight initialization strategies—may enable deeper models to capture subtle anatomical features while partially offsetting the trade-offs associated with increased model complexity [42].

# Conclusion

This study demonstrates that different architectural families for LVEF estimation exhibit distinct training dynamics and prediction behaviors, underscoring clear trade-offs between architectural complexity and generalization performance. Among the evaluated approaches, Inception-based architectures proved particularly effective, with more streamlined variants consistently achieving superior results. Notably, the Single-Input Fusion model outperformed the Dual-Input Fusion model, suggesting that *emergent feature specialization* may be more effective than *explicitly enforced stream separation* for this task especially in clinically critical low–EF regimes.

Across all architectures, overfitting emerged as the primary limiting factor, highlighting the importance of carefully balancing model capacity and regularization. Although none of the evaluated models surpassed the performance of EchoNet-Dynamic, our findings emphasize several design principles critical to effective LVEF estimation: the importance of sufficiently large effective receptive fields, avoidance of unnecessary architectural complexity, mitigation of vanishing and exploding gradients, and adequate model depth to capture meaningful spatiotemporal features.

The consistent difficulty in accurately identifying outlier LVEF cases further points to limitations inherent in the available data. Expanding the dataset to include a broader and more diverse range of echocardiographic presentations would likely enable the successful deployment of deeper and more expressive architectures, improving performance across both typical and clinically challenging cases. Addressing these data imbalances, alongside informed architectural design, is essential for advancing deep learning systems that not only match but potentially exceed human expertise in cardiac function assessment. More broadly, the insights gained from this work may also inform the design of robust models for other medical and non-medical video-based prediction tasks.